\documentclass{article} 
\usepackage{iclr2027_conference,times}

\usepackage{amsmath,amsfonts,bm}

\def\eqref#1{equation~\ref{#1}}

\def\1{\bm{1}}

\DeclareMathAlphabet{\mathsfit}{\encodingdefault}{\sfdefault}{m}{sl}
\SetMathAlphabet{\mathsfit}{bold}{\encodingdefault}{\sfdefault}{bx}{n}

\usepackage{hyperref}
\usepackage{url}
\usepackage{booktabs}
\usepackage{multirow}
\usepackage[table]{xcolor}
\usepackage{graphicx}
\usepackage{bbm}
\usepackage{mathtools}
\usepackage{enumitem}
\usepackage{hyperref}
\usepackage{cleveref}
\usepackage{wrapfig}
\usepackage[most]{tcolorbox}
\usepackage{listings}
\usepackage[T1]{fontenc}
\usepackage{textcomp}
\definecolor{promptback}{RGB}{247,248,250}
\definecolor{promptframe}{RGB}{86,96,110}

\newtcblisting{promptbox}[2][]{%
  enhanced,
  breakable,
  listing only,
  colback=promptback,
  colframe=promptframe,
  boxrule=0.5pt,
  arc=1mm,
  left=1.5mm,
  right=1.5mm,
  top=1mm,
  bottom=1mm,
  title={#2},
  fonttitle=\bfseries\small,
  listing options={%
    basicstyle=\ttfamily\footnotesize,
    breaklines=true,
    breakatwhitespace=false,
    columns=fullflexible,
    keepspaces=true,
    showstringspaces=false,
    upquote=true
  },
  #1
}

\newcommand{\ourmodel}{PILOT}
\crefformat{section}{\S#2#1#3}
\crefformat{subsection}{\S#2#1#3}
\crefformat{subsubsection}{\S#2#1#3}
\crefformat{appendix}{Appx. #2#1#3}
\crefformat{equation}{Eq.~(#2#1#3)}
\crefformat{table}{Tab.~#2#1#3}
\crefformat{figure}{Fig.~#2#1#3}
\crefformat{algorithm}{Alg.~#2#1#3}

\definecolor{pilotgainblue}{RGB}{30,80,160}
\definecolor{pilotrowblue}{RGB}{235,241,252}
\newcommand{\abldiff}[1]{\,{\scriptsize\textcolor{gray}{($#1$)}}}

\newcommand{\pilotgain}[2]{%
  #1\,{\scriptsize\textcolor{pilotgainblue}{\textbf{(#2)}}}%
}

\title{Beyond Post-Hoc Rewards: Privileged Reasoning Exploration for Multimodal Embeddings}

\author{Tianyu Yang$^{1}$, Yijun Tian$^{2}$\thanks{Corresponding author. \textsuperscript{\textdagger}Shared last authorship}, Lars Benedikt Kaesberg$^{1}$, Jan Philip Wahle$^{1}$,  \\
\textbf{Bela Gipp}$^{1,\dagger}$, \textbf{Terry Ruas}$^{1,\dagger}$, \textbf{Daniel Kurzawe}$^{1,3,\dagger}$    \\
$^{1}$University of Göttingen \quad $^{2}$University of Notre Dame \\ $^{3}$State and University Library of Göttingen  \\
\texttt{tianyu.yang@uni-goettingen.de, meetyijun@gmail.com} \\
}

\iclrfinalcopy 
\begin{document}

\maketitle

\begin{abstract}
Advances in Multimodal Large Language Models (MLLMs) have driven substantial progress in multimodal embeddings. 
Recent methods exploit generative capabilities by augmenting embedded inputs with self-generated reasoning to produce more informative embeddings.
However, conventional retrieval supervision for training embedding models specifies pairwise relevance relations without explicitly guiding the reasoning used for input augmentation.
Such reasoning is neither directly observed nor uniquely determined by the input, allowing multiple plausible traces to introduce semantic cues of varying retrieval utility.
Existing approaches largely optimize reasoning conditioned on the input alone, limiting the exploration of retrieval-relevant traces.
We introduce \textbf{PILOT}, a framework that models reasoning as a latent variable and instantiates its prior and posterior distributions as distinct reasoning modes of a shared MLLM.
The prior conditions only on the input, whereas the posterior uses additional training-time relational evidence to guide reasoning exploration.
To translate the posterior's broader exploration into inference-time improvements, we derive an importance-weighted variational objective and instantiate it as a prior–posterior co-evolution framework that improves both modes under a shared inference-compatible reward while selectively transferring posterior trajectories to the prior according to their utility and prior reachability.
On the 78-task MMEB-V2 benchmark, \ourmodel~outperforms the primary baseline across all 12 task groups at both 2B and 7B parameters, improving overall scores by 3.2 and 3.3 points, respectively, with the largest gains in visual-document retrieval (+6.1 and +6.4 points).
Our results show that relational supervision is most valuable not as a post-hoc reward but as a signal for shaping how reasoning is explored in training reasoning-enhanced embedding models.
\end{abstract}

\section{Introduction}
\label{sec:introduction}
Multimodal embeddings map heterogeneous inputs such as text, images, and videos into a shared dense representation space, supporting applications from cross-modal retrieval to clustering and recommendation~\citep{zhai2022lit, frome2013devise, 10.1145/3343031.3351034, 10.1145/3503161.3547755}.
Recent advances in multimodal large language models (MLLMs)~\citep{singh2025openai, comanici2025gemini, bai2025qwen3, wang2025internvl3} have shifted the field beyond the dual-encoder vision–language models ~\citep{radford2021learning, jia2021scaling} toward instruction-aware universal multimodal embeddings (UME) models, which follow nuanced instructions across tasks such as visual question answering and video moment retrieval~\citep{meng2026vlmvecv}, and generalize better in zero-shot settings~\citep{jiang2025vlmvec, meng2026vlmvecv, 10.1145/3746027.3754845, lan-etal-2025-llave}.
Yet most UME models still use the MLLM purely as an encoder, leaving its generative reasoning capacity untapped.

Chain-of-thought (CoT) reasoning improves MLLMs performance across diverse multimodal tasks by enabling a deeper understanding of complex inputs~\citep{zhang2024multimodal}.
Several works extend this idea to UME by augmenting each query or candidate with a self-generated reasoning trace and encoding the input together with its trace into a single embedding~\citep{cui2026think, lan2026ume, sun2026grace}.
Despite their effectiveness, existing approaches face a fundamental mismatch between reasoning generation and retrieval supervision. 
During training, queries are paired with relevant candidates with other candidates serving as negatives. 
These positive–negative relations identify relevant matches but leave the reasoning needed to distinguish them unspecified.
Consequently, multiple reasoning traces may be relevant to the same input yet differ substantially in their ability to distinguish positive candidates from negative ones.
Existing methods sample reasoning from the individual inputs alone, with relational supervision applied only through post-hoc rewards.
This creates a fundamental exploration bottleneck. Without access to relational evidence, the model is unlikely to discover the discriminative reasoning needed to distinguish relevant candidates from distractors, and post-hoc rewards can only reinforce already sampled sub-optimal trajectories.

\begin{figure}[t]
\begin{center}
\includegraphics[width=1.0\textwidth]{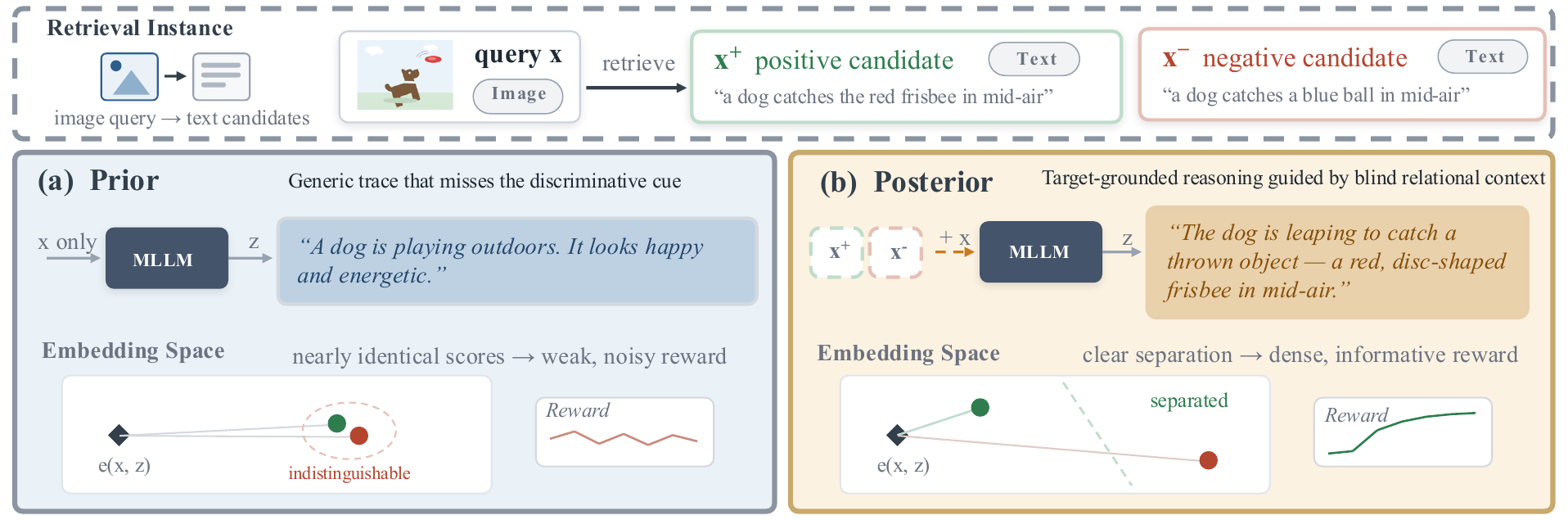}
\end{center}
\vspace{-10pt}
\caption{
\textbf{Relationally guided reasoning exploration in PILOT}. (a) Input-only prior reasoning may miss discriminative cues, yielding weak embedding separation. (b) The posterior uses training-time relational evidence to discover target-grounded cues that help distinguish positive and negative candidates. Embedding layouts and reward curves are schematic.
}
\vspace{-10pt}
\end{figure}

We address this bottleneck with \textbf{PILOT} (Privileged-Information-guided Latent Reasoning Optimization and Transfer), a framework that models reasoning traces as latent variables.
This formulation naturally distinguishes an inference-time \textbf{prior} conditioned only on the target input from a training-time privileged \textbf{posterior} that additionally uses relational evidence to infer retrieval-relevant reasoning.
We instantiate both distributions as distinct reasoning modes of a single MLLM.
Explicitly identifying the positive and negative counterparts allows the model to explain a given relevance relation without first determining which counterpart better matches the target.
To encourage more fine-grained, discriminative reasoning, we construct a \emph{blind contrastive posterior} that observes the target alongside a randomly ordered positive--negative pair with relevance labels withheld.
The model must compare both candidates against the target and predict which is positive, encouraging it to identify subtle semantic distinctions that determine relevance.
Using mode-specific prompt templates, a stronger annotation MLLM synthesizes reasoning traces for both distributions. 
Posterior traces are retained only when the annotator correctly identifies the positive counterpart.
The prior traces and verified posterior traces jointly initialize the two modes in the shared embedding MLLM through supervised fine-tuning (SFT).

Although the privileged posterior broadens the exploration of retrieval-relevant reasoning, only the input-conditioned prior is available at inference.
Moreover, not every posterior trajectory is useful or reproducible from the target input alone, making indiscriminate distillation undesirable.
We therefore derive an importance-weighted autoencoding (iWAE) objective~\citep{burda2015importance} with a mixture of the prior and posterior as the proposal distribution.
Its gradient naturally decomposes into on-policy improvement of the prior, on-policy improvement of the posterior, and importance-weighted transfer into the prior.
We instantiate these signals in a prior–posterior co-evolution procedure that uses GRPO~\citep{shao2024deepseekmath} for the two policy-improvement terms and weighted SFT for posterior-to-prior transfer.
Together, these updates allow privileged and inference-time reasoning to co-evolve under a unified objective, with downstream task reward and prior reachability governing which trajectories are exchanged rather than imposing a fixed teacher–student hierarchy.

We evaluate our overall approach on MMEB-v2~\citep{meng2026vlmvecv}. 
\ourmodel~outperforms the primary baseline across all 12 task groups at both 2B and 7B scales, improving overall scores by 3.2 and 3.3 points, respectively, with the largest gains in visual-document retrieval.
Ablations studies further demonstrate the benefits of dual-mode initialization, adaptive posterior exploration, and importance-weighted transfer.

Our main contributions can be summarized as follows:
\begin{itemize}[leftmargin=*, topsep=-0.5pt, itemsep=-0.5pt]
    \item We identify a fundamental exploration bottleneck in reasoning-enhanced multimodal embeddings. Existing methods generate reasoning from individual inputs while using positive–negative relations only for post-hoc evaluation, limiting the discovery of discriminative traces.
    \item We introduce \textbf{\ourmodel}, a framework that models reasoning as a latent variable, pairing an input-conditioned prior for inference with a privileged posterior constructed from blind relational evidence during training.
    \item We derive a mixture importance-weighted variational objective that motivates prior--posterior co-evolution, combining joint policy improvement with selective transfer to the prior based on retrieval utility and prior reachability.
    \item Experiments on the 78-task MMEB-V2 benchmark show that \ourmodel~outperforms primary multimodal embedding baselines at both 2B and 7B scales, highlighting the value of relational supervision not only for evaluating reasoning, but also for guiding its exploration and selective transfer to the inference-time prior.
\end{itemize}

\section{Related Work}
\label{sec:related_work}

\subsection{Universal Multimodal Embedding}
\label{subsec:universal_multimodal_embedding}
Recent methods adapt MLLMs into UME models through contrastive learning and hard-negative mining, but typically encode inputs directly~\citep{jiang2025vlmvec, meng2026vlmvecv, lin2025mmembed}.
Think-Then-Embed~\citep{cui2026think} and Embed-RL~\citep{jiang2026embed} introduce separate reasoner--embedder pipelines with the embedder encoding each input together with a reasoning trace generated by the reasoner.
UME-R1 unifies reasoning generation and embedding extraction within a single model trained with SFT and RL, while PLUME replaces explicit textual traces with compact continuous reasoning~\citep{lan2026ume,he2026plumelatentreasoningbased}.
MMEmb-R1 and UniME-R1 further incorporate relational signals through pair-aware trace selection and retrieval-centric reasoning construction~\citep{wang2026mmemb,sun2026learning}.
These signals primarily evaluate, select, or refine reasoning after generation, whereas our privileged posterior uses training-time relational evidence to directly guide trajectory exploration while preserving input-only inference.

\subsection{Learning with Privileged Information}
\label{subsec:learning_with_privileged_information}
Privileged information refers to auxiliary evidence available during training that can guide learning but is unavailable to the deployed model at inference~\citep{2789272.2886814}.
Prior work in language-model reasoning uses such information either to distill a fixed privileged teacher into a deployable model or to jointly and iteratively improve privileged and deployable reasoning models~\citep{agarwal2024onpolicy,penaloza2026privileged,lin-etal-2026-ravr,liang2026search,zhou2026variational,wen2026coupled}.
However, these methods primarily address answer-generation tasks, where privileged information reveals or supports the reconstruction of a target output.
In multimodal embedding, a relevant counterpart is not an output to reconstruct but a separate input whose relation to the input, relative to negatives, defines the supervision.
This difference changes both how relational evidence should condition privileged reasoning and how reasoning utility should enter the learning objective.
We address this setting by constructing a privileged posterior from positive–negative relations and co-evolving it with the input-conditioned prior under a unified latent variable objective.

\section{Methodology}
\label{sec:methodology}

\begin{figure}[t]
\begin{center}
\includegraphics[width=1.0\textwidth]{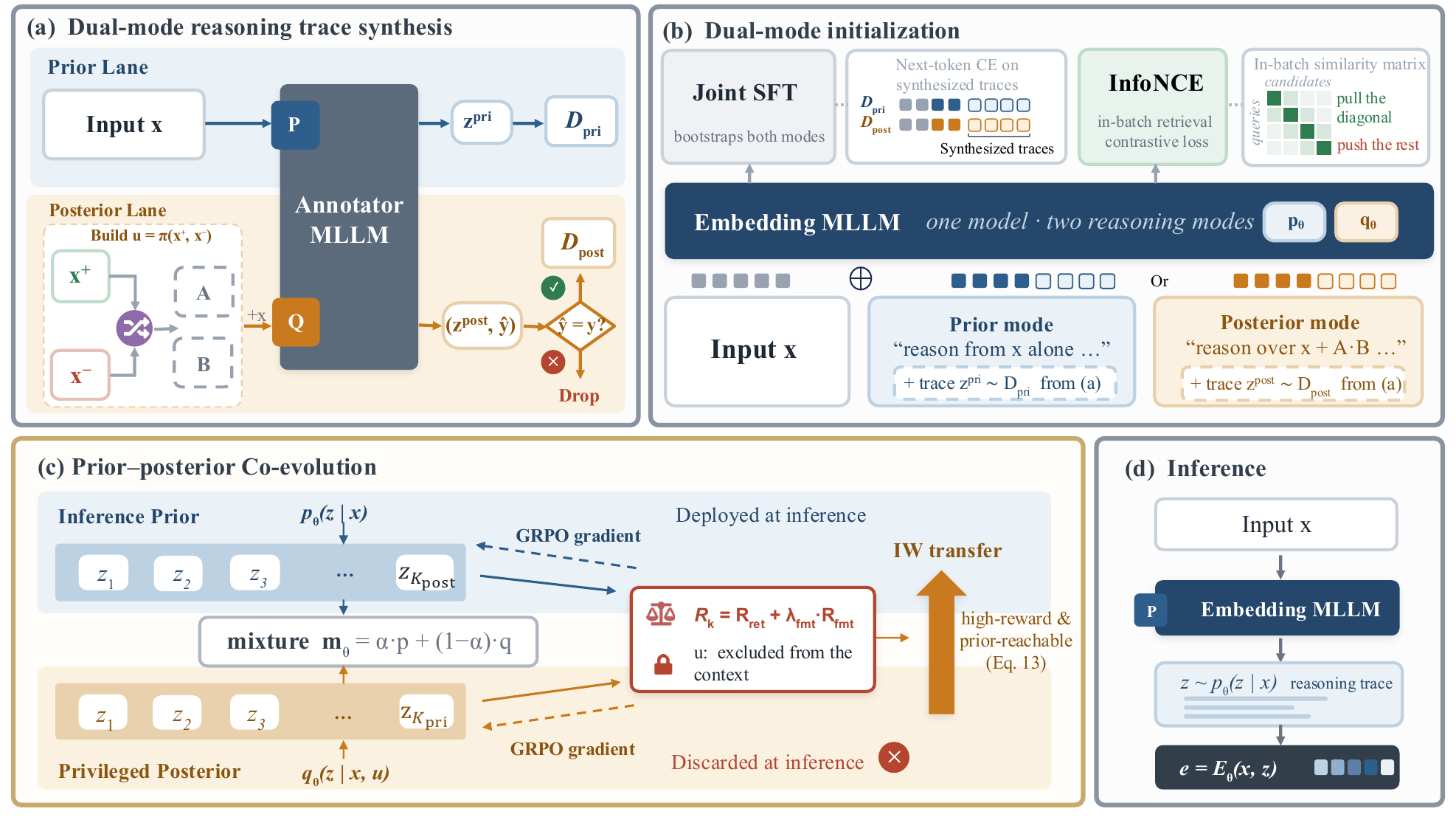}
\end{center}
\vspace{-10pt}
\caption{\textbf{Overview of \ourmodel.} \textbf{(a)}~A stronger offline annotator, given only a shuffled, unlabeled candidate pair $u$, synthesizes prior and consistency-filtered posterior reasoning traces. \textbf{(b)}~Joint SFT and contrastive learning initialize both reasoning modes in a shared MLLM (Eq.~\ref{eq:joint_init}). \textbf{(c)}~GRPO optimizes both modes using rewards evaluated under the prior context (Eq.~\ref{eq:inference_compatible_reward}), while importance-weighted fitting transfers trajectories to the prior according to retrieval utility and prior reachability (Eq.~\ref{eq:transfer}). \textbf{(d)}~Inference runs the prior alone.}
\vspace{-10pt}
\end{figure}

\subsection{Problem Formulation}
\label{sec:latent_reasoning}
We consider multimodal retrieval where queries and candidates may contain text, images, videos, or interleaved multimodal content.
Let $x$ denote a target input to be embedded, which can be either a query or a candidate.
Its observed retrieval relation is denoted by $\mathcal R=(x^{+},\mathcal X^{-})$, where $x^{+}$ is a positive counterpart and $\mathcal X^{-}$ is a set of negatives. 
Following the unified architecture of UME-R1, the model first samples a reasoning trace $z$ and then encodes it together with the target input:
\begin{equation}
z \sim p_\theta(z\mid x),
\qquad
\mathbf e_\theta(x,z)=E_\theta(x,z).
\label{eq:reasoning_embedding}
\end{equation}
Here, $\mathbf e_\theta(x,z)\in\mathbb R^d$ is the normalized embedding, and $E_\theta$ extracts and normalizes the final hidden state obtained by encoding $x$ together with $z$.
Retrieval supervision specifies $\mathcal R$ but does not provide a corresponding reasoning trace. Moreover, multiple traces may be plausible for the same input while emphasizing different semantics and yielding substantially different retrieval outcomes.
We therefore propose to treat $z$ as a latent variable and marginalize over possible reasoning traces:
\begin{equation}
p_{\theta}(\mathcal{R}\mid x)
=
\sum_{z}
p_{\theta}(z\mid x)
p(\mathcal{R}\mid x,z).
\label{eq:latent_retrieval_likelihood}
\end{equation}
Here, $p_{\theta}(z\mid x)$ serves as the reasoning \emph{prior}. It generates each trace from its corresponding input alone, matching the independent encoding procedure required at inference time.
We connect the conditional likelihood to the downstream retrieval reward through
\begin{equation}
p(\mathcal{R}\mid x, z)
\propto
\exp\left(
\beta R_{\mathrm{ret}}(\mathbf{e}_\theta(x,z),\mathcal R)
\right),
\label{eq:reward_likelihood}
\end{equation}
where $R_{\mathrm{ret}}$ evaluates the embedding induced by $z$ and $\beta$ controls the concentration on high-utility trajectories. 
Equation~\ref{eq:latent_retrieval_likelihood} exposes the exploration bottleneck of existing methods.
In practice the sum is approximated with a few traces from the input-only prior, and retrieval rewards can only score these after the fact. Reasoning that separates the positive from plausible negatives receives no signal unless it is sampled, yet it must be discovered without ever observing those candidates.
Conditioning on the observed relation induces an ideal posterior distribution over the reasoning trace:
\begin{equation}
p_{\theta}^{\star}(z\mid x,\mathcal{R})
=
\frac{
p_{\theta}(z\mid x)
p(\mathcal{R}\mid x,z)
}{
p_{\theta}(\mathcal{R}\mid x)
},
\label{eq:ideal_reasoning_posterior}
\end{equation}
which reweights the input-only prior according to the retrieval utility of each reasoning trajectory.
This posterior therefore places greater probability on reasoning trajectories that explain the observed retrieval relation and that the prior should ultimately learn to recover.
However, it is intractable to compute exactly for sampling. 
We therefore introduce a privileged variational \emph{posterior} $q_\theta(z\mid x,u)$ where $u$ denotes the relational context associated with $\mathcal{R}$.
This posterior serves as an informed proposal for exploring reasoning trajectories that are difficult to discover from individual inputs alone.
Both distributions are instantiated by the same MLLMs under different conditioning prompts.

\subsection{Dual-Mode Initialization}
\label{sec:dual_mode_initialization}
The latent-variable formulation motivates a privileged posterior that leverages retrieval relations to explore reasoning trajectories less likely under the input-only prior.
To elicit more discriminative reasoning, we condition it on randomly ordered, unlabeled positive and negative counterparts, yielding a \emph{blind contrastive posterior}.
For each target input $x$, we sample its positive counterpart $x^{+}$ and a negative counterpart $x^{-}\in\mathcal{X}^{-}$, randomly permute them, and assign anonymous identifiers:
\begin{equation}
u=(c_A,c_B)=\pi(x^{+},x^{-}),
\qquad
y\in\{A,B\},
\label{eq:blind_evidence}
\end{equation}
where $\pi$ is a random permutation and $y$ denotes the hidden position of the positive candidate. The privileged posterior observes the target and the unlabeled candidate pair, but not $y$:
\begin{equation}
z^{\mathrm{post}}
\sim
q_\theta(z\mid x,u).
\label{eq:blind_posterior}
\end{equation}
By withholding relevance labels and randomizing candidate order, this construction encourages the model to compare the counterpart against the input and identify the semantic distinctions that determine relevance, eliciting more discriminative reasoning.
However, compact embedding MLLMs may struggle to generate such reasoning in a zero-shot setting.
Indiscriminately distilling privileged traces into $p_\theta(z\mid x)$ ignores privileged traces’ varying retrieval utility and prior reachability.
We thus synthesize traces offline with a stronger model $\mathcal{A}$ and initialize both reasoning modes through SFT.
Conditioned only on the target input $x$, $\mathcal{A}$ generates a prior trace $z^{\mathrm{pri}}$. Conditioned on $x$ and the blind contrastive context $u$, it jointly generates a posterior trace and predicts the positive candidate:
\begin{equation}
(z^{\mathrm{post}},\hat y)
\sim
\mathcal{A}(\cdot\mid x,u).
\label{eq:posterior_synthesis}
\end{equation}
We retain a synthesized posterior annotation only if $\hat y=y.$ 
This consistency check filters traces generated from incorrect candidate comparisons or hallucinated relations, ensuring that retained posterior traces reflect a correct interpretation of the retrieval supervision.

Finally, We use the synthesized prior and verified posterior traces $\mathcal D_{\mathrm{pri}}\cup\mathcal D_{\mathrm{post}}$ to initialize the embedding model with a joint generation-and-retrieval objective:
\begin{equation}
     \mathcal{L}_{\mathrm{init}}=-\mathbb{E}_{\mathcal{D}_{\mathrm{pri}}}\log p_\theta({z}^{\mathrm{pri}}|x) - \mathbb{E}_{\mathcal{D}_{\mathrm{post}}}\log q_\theta ({z}^{\mathrm{post}}\mid x,{u})+\lambda_{\mathrm{ret}}\ell_{\mathrm{NCE}}\bigl(\mathbf{E}_{\theta}(\mathcal{B})\bigr)
\label{eq:joint_init}
\end{equation}
where $\mathbf{E}_{\theta}(\mathcal{B})$ denotes the collection of reasoning-conditioned query and candidate embeddings in minibatch $\mathcal{B}$, $\ell_{\mathrm{NCE}}(\cdot)$ is the standard in-batch contrastive loss and $\lambda_{\mathrm{ret}}$ controls its weight.
The first two terms bootstrap the inference-time prior and privileged posterior mode, respectively, while the InfoNCE loss aligns the reasoning-augmented representations with retrieval discrimination.

\subsection{Prior--Posterior Co-Evolution}
\label{sec:\ourmodel}
Dual-mode initialization establishes the prior and privileged posterior, but leaves open how privileged exploration should improve inference-time reasoning. 
We therefore develop a prior–posterior co-evolution procedure that jointly optimizes both reasoning modes with retrieval rewards and couples them through importance-weighted prior fitting.

To combine inference-compatible reasoning with privileged exploration, we construct the variational proposal as a mixture of the prior and posterior:
\begin{equation}
    m_{\theta}(z\mid x,{u})
    =
    \alpha p_{\theta}(z\mid x)
    +
    (1-\alpha)q_{\theta}(z\mid x,{u}),
    \label{eq:mixture_proposal}
\end{equation}
where $\alpha\in(0,1)$ determines the proportion of prior rollouts in the overall proposals.
In practice, we draw $K_{\mathrm{pri}}$ trajectories from the prior and $K_{\mathrm{post}}$ trajectories from the posterior, with $\alpha=K_{\mathrm{pri}}/(K_{\mathrm{pri}}+K_{\mathrm{post}})$.
Using the reward-induced likelihood introduced in Equation~\ref{eq:reward_likelihood}, the latent evidence can be estimated with $K=K_{\mathrm{pri}}+K_{\mathrm{post}}$ samples from the mixture proposal through the importance-weighted lower bound:
\begin{equation}
\begin{aligned}
    \mathcal{J}_{IW}
    &=
    \mathbb{E}_{z_{1:K}\sim m_{\theta}}
    \left[
        \log
        \frac{1}{K}
        \sum_{k=1}^{K} w_k
    \right]
    \leq
    \log p_{\theta}(\mathcal{R}\mid x),
    \\
    w_k
    &=
    \frac{
        p_{\theta}(z_k\mid x)
        p(\mathcal{R}\mid x,z_k)
    }{
        m_{\theta}(z_k\mid x,u)
    }
    \propto
    \frac{
        p_{\theta}(z_k\mid x)
        \exp\left({\beta R_{\mathrm{ret}}(x,z_k)}\right)
    }{
        m_{\theta}(z_k\mid x,u)
    }.
\end{aligned}
\label{eq:\ourmodel_iwae}
\end{equation}
The importance weight combines two complementary criteria.
The reward term measures the retrieval utility of a trajectory, while the ratio $p_{\theta}(z_k\mid x)/m_{\theta}(z_k\mid x,u)$ measures its reachability under the inference-time prior.. 
Let $\bar{w}_k=\frac{w_k}{\sum_j w_j}$ denote the normalized importance weight. 
The gradient of Equation~\ref{eq:\ourmodel_iwae} separates into three types of learning signals:
\begin{equation}
    \nabla_\theta\mathcal{J}_{\mathrm{IW}}
=
\mathbb{E}_{p_\theta}\!\left[A^{p}\nabla_\theta\log p_\theta(z|x)\right]
+
\mathbb{E}_{q_\theta}\!\left[A^{q}\nabla_\theta\log q_\theta(z|x,u)\right]
+
\mathbb{E}_{m_\theta}\!\left[\bar w\nabla_\theta\log p_\theta(z|x)\right] .
\label{eq:gradient_decomposition}
\end{equation}
Detailed derivation can be seen in Appendix~\ref{app:mixture_iwae_derivation}. 
Here, $A^{p}$ and $A^{q}$ denote the score-function learning signals for prior- and posterior-sampled trajectories, respectively, and the expectations are taken over the prior, posterior, and mixture distributions. 
Samples originating from the prior improve the prior as an on-policy reasoning distribution, while samples originating from the posterior similarly improve privileged exploration. In addition, the prior likelihood in the numerator provides an importance-weighted learning signal that transfers useful mixture samples toward the prior.

Directly optimizing the full score-function gradient can cause high variance. We therefore approximate the two policy-improvement terms using GRPO. 
For every sampled trajectory, we compute an inference-compatible reward
\begin{equation}
R_k
=
R_{\mathrm{ret}}
\bigl(
\mathbf e_{\theta}(x,z_k),
\mathcal R
\bigr)
+
\lambda_{\mathrm{fmt}}R_{\mathrm{fmt}}(z_k).
\label{eq:inference_compatible_reward}
\end{equation}
where $R_{\mathrm{fmt}}$ denotes the format reward function and $\lambda_{\mathrm{fmt}}$ controls its effect.
Although posterior trajectories are generated using $u$, the privileged evidence is excluded when constructing $\mathbf e_{\theta}(x,z_k)$. Thus, privileged information can improve the reward only by inducing a more useful reasoning trace, rather than by directly entering the evaluated embedding.
Rewards are then normalized within each reasoning mode to obtain group-relative advantages, producing the losses
$\mathcal{L}_{\mathrm{GRPO}}^{\mathrm{pri}}$ and
$\mathcal{L}_{\mathrm{GRPO}}^{\mathrm{post}}$.
We instantiate the cross-distribution transfer term using weighted supervised fine-tuning and retain only the posterior-sampled component to avoid overly reinforcing the prior samples.
The objective is defined as:
\begin{equation}
    \mathcal{L}_{\mathrm{transfer}}
    =
    -
    \sum_{k:s_k=\mathrm{post}}
    \operatorname{sg}
    \left[
        \bar{w}_k
    \right]
    \log p_{\theta}(z_k\mid x),
\label{eq:transfer}
\end{equation}
where $\operatorname{sg}[\cdot]$ stops gradients through the transfer weights. Unlike conventional distillation, this objective does not require the prior to imitate every posterior trajectory. Posterior trajectories contribute strongly only when they achieve high reward and remain sufficiently reachable under the prior.

The final \ourmodel~objective is
\begin{equation}
    \mathcal{L}_{\mathrm{\ourmodel}}
    =
    \mathcal{L}_{\mathrm{GRPO}}^{\mathrm{pri}}
    +
    \lambda_{\mathrm{post}}
    \mathcal{L}_{\mathrm{GRPO}}^{\mathrm{post}}
    +
    \lambda_{\mathrm{transfer}}
    \mathcal{L}_{\mathrm{transfer}},
\end{equation}
where $\lambda_{\mathrm{post}}$ and $\lambda_{\mathrm{transfer}}$ control privileged policy improvement and posterior-to-prior transfer, respectively. This optimization allows the prior and posterior to co-evolve: the posterior uses relational evidence to explore retrieval-relevant reasoning, the prior improves directly under inference-compatible rewards, and importance weighting selectively internalizes privileged discoveries that are both useful and reproducible.

\section{Experiments}
\label{sec:experiments}

\subsection{Experimental Setup}
\label{sec:experimental_setup}

\paragraph{Dataset and metrics.}
We build on the datasets released by UME-R1, comprising 1.46M SFT examples and 11,136 RL examples curated from MMEB-Train~\citep{jiang2025vlmvec}.
For dual-mode initialization, we adapt the SFT dataset. 
We synthesize posterior traces and retain approximately 676K examples after label-prediction verification for posterior-mode supervision, while preserving the original traces of the remaining 787K examples annotated by GLM-4.1V-thinking~\citep{hong2025glm} for prior-mode supervision. 
Co-evolution uses the released RL dataset.
We evaluate \ourmodel~on the full MMEB-V2 benchmark~\citep{meng2026vlmvecv}, which comprises 78 tasks spanning image, video, and visual-document retrieval.
Following the standard protocol, we report Hit@1 for image and video tasks and nDCG@5~\citep{10.1145/582415.582418} for visual-document tasks.
In the analysis experiment, we additionally report Pass@K that measures the proportion of queries for which at least one of $K$ sampled reasoning traces produces an embedding that ranks an official positive candidate first.

\paragraph{Baselines.} Our primary baseline, UME-R1, shares \ourmodel’s underlying training data, Qwen2-VL~\citep{wang2024qwen2} backbones, and unified embedding setup.
DUME is its encoding-only variant without reasoning generation.
TTE$_s$ and MMEmb-R1 (reasoning-enhanced baselines) differ in data curation and reasoning supervision, with TTE$_s$ using separate reasoner and embedder models.
We also compare against GME~\citep{zhang2024gme}, VLM2Vec~\citep{jiang2025vlmvec}, VLM2Vec-V2~\citep{meng2026vlmvecv}, LamRA~\citep{liu2025lamra}, CAFe~\citep{11375171}, and RzenEmbed~\citep{jian2025rzenembed}, alongside the document-specialized multi-vector retriever ColPali-V1.3~\citep{faysse2025colpali} and proprietary Seed-1.6-Embedding. 
Results are grouped by backbone size, with differences in training data, reasoning supervision, and inference configurations taken into account when interpreting comparisons.

We further include DUME, a discriminative-only baseline trained under matched settings, and UME-R1~\citep{lan2026ume}, our closest baseline, which generates intermediate reasoning before embedding extraction and optimizes it through SFT and RL. Where available, we report both 2B and 7B variants and prioritize comparisons with models using the same Qwen2-VL backbone~\citep{wang2024qwen2} and parameter scale; CAFe and ColPali-V1.3 instead use LLaVA-OneVision~\cite{li2025llavaonevision} and PaliGemma~\citep{beyer2024paligemma}, respectively.
\paragraph{Implementation details.}
We use Qwen2-VL-2B/7B-Instruct backbones and GLM-4.1V-thinking for posterior trace synthesis.
Initialization runs for 5,000 steps with a global batch size of 1,024 via gradient accumulation, a learning rate of $5\times10^{-5}$, and a maximum sequence length of 12,288 tokens. 
Unless otherwise specified, co-evolution samples 8 trajectories per input per mode, with a global batch size of 64 and a learning rate of $10^{-6}$. For efficient evaluation, we modify vLLM~\citep{kwon2023efficient} to return the normalized hidden state at a designated special token after generation.
Additional implementation details are provided in Appendix~\ref{app:implementation_details}.

\subsection{Main Results}

\begin{table*}[t]
\caption{
Performance comparison across image, video, and visual-document (VisDoc)
retrieval benchmarks.
Shaded rows denote \ourmodel{}.
Blue values in parentheses indicate absolute gains over UME-R1
at the same backbone size. 
}
\label{tab:main_results}
\centering
\setlength{\tabcolsep}{3pt}
\renewcommand{\arraystretch}{1.12}

\resizebox{\textwidth}{!}{%
\begin{tabular}{lcc*{16}{c}}
\toprule
\multirow{2}{*}{\textbf{Model}}
& \multicolumn{2}{c}{\textbf{Training}}
& \multicolumn{5}{c}{\textbf{Image}}
& \multicolumn{5}{c}{\textbf{Video}}
& \multicolumn{5}{c}{\textbf{VisDoc}}
& \multirow{2}{*}{\textbf{All}} \\
\cmidrule(lr){2-3}
\cmidrule(lr){4-8}
\cmidrule(lr){9-13}
\cmidrule(lr){14-18}
& \# of Pairs & \# of Annotators
& CLS & QA & RET & GD & Avg.
& CLS & QA & RET & MRET & Avg.
& VDRv1 & VDRv2 & VR & OOD & Avg.
& \\
\midrule
\multicolumn{3}{l}{\textbf{\# of Datasets}}
& 10 & 10 & 12 & 4 & 36
& 5 & 5 & 5 & 3 & 18
& 10 & 4 & 6 & 4 & 24
& 78 \\
\midrule
\multicolumn{19}{c}{\textit{Closed-Source Models}} \\
\midrule

Seed-1.6-embedding
& NR & NR
& 76.1 & 74.0 & 77.9 & 91.3 & 77.8
& 55.0 & 60.9 & 51.3 & 53.5 & 55.3
& 85.5 & 56.6 & 84.7 & 43.1 & 73.4
& 71.3 \\

\midrule
\multicolumn{18}{c}{\textit{Small Backbones (2--3B)}} \\
\midrule

ColPali-V1.3
& 127K & ---
& 40.3 & 11.5 & 48.1 & 40.3 & 34.9
& 26.7 & 37.8 & 21.6 & 25.5 & 28.2
& 83.6 & 52.0 & 81.1 & 43.1 & 71.0
& 44.4 \\

VLM2Vec
& 1.03M$^{*}$ & ---
& 58.7 & 49.3 & 65.0 & 72.9 & 59.7
& 33.4 & 30.5 & 20.6 & 33.0 & 29.0
& 49.8 & 13.5 & 51.8 & 33.5 & 41.6
& 47.0 \\

DUME
& 1.46M & ---
& 59.3 & 55.0 & 66.3 & 78.0 & 62.5
& 37.7 & 46.6 & 17.1 & 30.0 & 33.2
& 67.6 & 43.3 & 47.1 & 33.8 & 52.8
& 52.7 \\

GME
& 8M & ---
& 54.4 & 29.9 & 66.9 & 55.5 & 51.9
& 34.9 & 42.0 & 25.6 & 32.4 & 33.9
& 86.1 & 54.0 & 82.5 & 43.1 & 72.7
& 54.1 \\

VLM2Vec-V2
& 2M$^{*}$ & ---
& 62.9 & 56.3 & 69.5 & 77.3 & 64.9
& 39.3 & 34.3 & 28.8 & 38.5 & 34.9
& 75.5 & 44.9 & 79.4 & 39.4 & 65.4
& 58.0 \\

RzenEmbed
& $\geq$5M & ---
& 68.5 & 66.3 & 74.5 & 90.3 & 72.3
& 50.4 & 49.7 & 46.6 & 38.9 & 47.3
& 87.1 & 55.1 & 87.2 & 43.4 & 74.5
& 67.2 \\

\midrule

TTE$_s^{\dagger}$
& 2M$^{*}$ & 1
& 67.9 & 66.6 & 70.2 & 84.1 & 70.1
& 47.3 & 49.1 & 34.4 & 33.2 & 41.9
& 77.5 & 53.2 & 83.2 & 41.1 & 68.8
& 63.1 \\

MMEmb-R1
& 1.2M & 3
& 64.5 & 68.1 & 70.0 & 88.9 & 70.0
& 56.3 & 52.8 & 47.6 & 42.5 & 50.6
& 74.1 & 56.0 & 78.9 & 48.3 & 68.0
& 65.0 \\

\addlinespace[3pt]

UME-R1
& 1.46M & 1
& 64.4 & 62.0 & 67.8 & 75.7 & 66.1
& 44.3 & 50.0 & 32.9 & 39.3 & 41.8
& 72.4 & 46.0 & 79.2 & 37.2 & 63.8
& 60.1 \\

\rowcolor{pilotrowblue}
\textbf{\ourmodel}
& 1.46M & 1
& 65.2 & 63.8 & 69.0 & 78.1
& \pilotgain{67.5}{+1.4}
& 47.7 & 58.3 & 34.8 & 40.8
& \pilotgain{45.9}{+4.1}
& 80.6 & 51.9 & 82.5 & 42.1
& \pilotgain{69.9}{+6.1}
& \pilotgain{63.3}{+3.2} \\

\midrule
\multicolumn{18}{c}{\textit{Large Backbones (\(\sim\) 7B)}} \\
\midrule
LamRA
& 1.38M$^{*}$ & ---
& 59.2 & 26.5 & 70.0 & 62.7 & 54.1
& 39.3 & 42.6 & 24.3 & 34.6 & 35.2
& 22.0 & 11.5 & 37.4 & 21.0 & 23.9
& 40.4 \\

GME
& 8M & ---
& 57.7 & 34.7 & 71.2 & 59.3 & 56.0
& 37.4 & 50.4 & 28.4 & 38.2 & 38.6
& 89.4 & 55.6 & 85.0 & 44.4 & 75.2
& 57.8 \\

VLM2Vec
& 1.03M$^{*}$ & ---
& 62.7 & 56.9 & 69.4 & 82.2 & 65.5
& 39.1 & 30.0 & 29.0 & 40.6 & 34.0
& 56.9 & 9.4 & 59.1 & 38.1 & 46.4
& 52.3 \\

DUME
& 1.46M & ---
& 64.2 & 57.0 & 70.8 & 81.8 & 66.4
& 32.9 & 47.4 & 8.6 & 28.0 & 29.4
& 67.1 & 35.2 & 82.6 & 34.9 & 60.3
& 55.9 \\

CAFe
& NR & ---
& 63.6 & 61.7 & 69.1 & 87.6 & 67.6
& 35.8 & 58.7 & 34.4 & 39.5 & 42.4
& 70.7 & 49.6 & 79.5 & 38.1 & 63.9
& 60.6 \\

VLM2Vec-V2
& 2M$^{*}$ & ---
& 65.7 & 61.5 & 70.0 & 85.2 & 68.1
& 45.9 & 33.9 & 27.6 & 39.3 & 36.4
& 78.8 & 52.6 & 82.7 & 42.1 & 69.3
& 61.2 \\

RzenEmbed
& $\geq$5M & ---
& 70.6 & 71.7 & 78.5 & 92.1 & 75.9
& 58.8 & 63.5 & 51.0 & 45.5 & 55.7
& 89.7 & 60.7 & 88.7 & 44.4 & 77.1 
& 71.6 \\ 

\midrule

TTE$_s^{\dagger}$
& 2M$^{*}$ & 1
& 69.7 & 72.4 & 74.0 & 90.6 & 74.2
& 49.1 & 60.6 & 36.4 & 37.2 & 46.8
& 84.1 & 62.7 & 91.9 & 47.6 & 76.4
& 68.6 \\

MMEmb-R1
& 1.2M & 3
& 68.1 & 71.8 & 77.7 & 87.1 & 74.4
& 60.2 & 56.0 & 51.2 & 41.1 & 53.3
& 79.6 & 65.3 & 84.7 & 58.0 & 74.9
& 69.7 \\

\addlinespace[3pt]

UME-R1
& 1.46M & 1
& 66.4 & 69.1 & 71.8 & 84.1 & 70.9
& 48.2 & 60.4 & 38.5 & 38.9 & 47.3
& 75.8 & 50.0 & 83.7 & 37.5 & 67.1
& 64.3 \\

\rowcolor{pilotrowblue}
\textbf{\ourmodel}
& 1.46M & 1
& 68.7 & 70.2 & 73.2 & 87.2
& \pilotgain{72.7}{+1.8}
& 50.1 & 62.8 & 40.2 & 41.4
& \pilotgain{49.4}{+2.1}
& 83.6 & 60.8 & 85.1 & 43.3
& \pilotgain{73.5}{+6.4}
& \pilotgain{67.6}{+3.3} \\

\bottomrule
\end{tabular}%
}

\smallskip
\begin{minipage}{\textwidth}
\footnotesize
$\dagger$ TTE$_s$ uses separate reasoner and embedder models.
--- indicates no trace supervision, NR an unknown total,
and $^{*}$ an estimate from released data.
Within each size group, \textbf{encoding-only} models precede the
internal rule and \textbf{reasoning-enhanced} models follow it.
\end{minipage}
\vspace{-10pt}
\end{table*}

Table~\ref{tab:main_results} shows that \ourmodel~outperforms UME-R1 across all 12 task groups at both scales, reaching overall scores of 63.3 (2B) and 67.6 (7B) using only the inference-time prior.
Tasks in the VisDoc domain benefits most, with gains of 6.1 and 6.4 points. Our 2B model even outperforms UME-R1-7B in this modality (69.9 vs 67.1), suggesting that improved training can partly compensate for a smaller backbone.
These gains over an already reasoning-enhanced baseline highlight the importance of how reasoning is learned for embedding.
Against encoding-only methods, \ourmodel~ surpasses VLM2Vec-V2 by 5.3 and 6.4 points overall and DUME by 10.6 and 11.7 points at the two scales. 
GME nevertheless achieves higher score in the VisDoc domain, while RzenEmbed performs better overall, indicating that the effectiveness of the complete training recipe remains important.
Among reasoning-enhanced methods, \ourmodel~performs comparably to TTE$_s$ at 2B (63.3 vs 63.1) and achieves a higher video average at 7B (49.4 vs 46.8), although TTE$_s$ and MMEmb-R1 score higher overall at 7B. 
These results reveal task-dependent strengths across different supervision and inference configurations, while the consistent gains over UME-R1 support using training-time relational evidence to improve inference-time embeddings.

\subsection{Ablation Study}

\begin{table*}[t]
\centering
\caption{Component ablation on MMEB-V2.
Each row specifies a complete training configuration.
All variants are evaluated using only the inference-time prior.}
\label{tab:ablation}

\setlength{\tabcolsep}{3pt}
\renewcommand{\arraystretch}{1.15}

\resizebox{\textwidth}{!}{%
\begin{tabular}{l*{9}{c}}
\toprule

& \multicolumn{5}{c}{\textbf{Training configuration}}
& \multicolumn{4}{c}{\textbf{Performance}} \\
\cmidrule(lr){2-6}
\cmidrule(lr){7-10}

\textbf{Variant}
& \shortstack{\textbf{SFT}\\\textbf{initialization}}
& \shortstack{\textbf{Prior}\\\textbf{GRPO}}
& \shortstack{\textbf{Posterior}\\\textbf{during RL}}
& \shortstack{\textbf{IW prior}\\\textbf{fitting}}
& \shortstack{\textbf{Rollouts/query}\\\textbf{(prior + posterior)}}
& \textbf{Image}
& \textbf{Video}
& \textbf{VisDoc}
& \textbf{All} \\
\midrule

\rowcolor{pilotrowblue}
\textbf{PILOT}
& Dual-mode
& $\checkmark$
& GRPO-updated
& $\checkmark$
& $8 + 8$
& 67.5 & 45.9 & 69.9 & 63.3 \\

Joint GRPO
& Dual-mode
& $\checkmark$
& GRPO-updated
& $\times$
& $8 + 8$
& 66.8 & 45.1 & 68.1 & 62.2\abldiff{-1.1} \\

Frozen-posterior variant
& Dual-mode
& $\checkmark$
& Frozen
& $\times$
& $8 + 0$
& 67.1 & 44.8 & 66.1 & 61.6\abldiff{-1.7} \\

Prior GRPO only
& Dual-mode
& $\checkmark$
& Not used
& $\times$
& $16 + 0$
& 66.5 & 43.4 & 67.2 & 61.4\abldiff{-1.9} \\

\midrule

Dual-mode SFT
& Dual-mode
& --- & --- & --- & ---
& 65.7 & 42.1 & 65.9 & 60.3\abldiff{-3.0} \\

Direct Distillation
& Prior
& --- & --- & --- & ---
& 64.7 & 41.6 & 64.8 & 59.4\abldiff{-3.9} \\

Prior SFT
& Prior
& --- & --- & --- & ---
& 65.3 & 41.0 & 63.2 & 59.0\abldiff{-4.3} \\

\bottomrule
\end{tabular}%
}

\smallskip
\begin{minipage}{\textwidth}
\footnotesize
$\checkmark$: enabled;
$\times$: disabled;
---: not applicable.
IW denotes importance weighting.
Rollouts/query reports the number of trajectories sampled
per query during RL, written as prior + posterior.
\end{minipage}
\vspace{-10pt}
\end{table*}

Table~\ref{tab:ablation} reports ablations using Qwen2-VL-2B-Instruct, with results averaged over three random seeds.
Dual-mode SFT improves the overall score from 59.0 to 60.3, with gains across all three modalities.
Direct distillation, which pairs input-only prior prompts with annotated posterior traces as SFT targets, achieves only 59.4.
Its limited gain is consistent with a conditioning mismatch, as posterior traces may depend on candidate information unavailable to the prior.
Prior-only GRPO further raises the score to 61.4, confirming benefits from both privileged initialization and direct optimization of the deployed reasoning mode.
Adding a frozen posterior yields only 0.2 points overall and reduces visual-document performance from 67.2 to 66.1. Joint GRPO instead reaches 62.2 overall and 68.1 on visual documents, outperforming prior-only GRPO across all modalities. 
This supports adapting the posterior during training rather than relying on fixed privileged guidance.
Importance-weighted prior fitting adds another 1.1 points overall, improving image, video, and visual-document scores by 0.7, 0.8, and 1.8 points, respectively. Full \ourmodel~exceeds prior-only GRPO by 1.9 points, showing that explicit transfer complements joint policy optimization.

\subsection{Further Analysis}
To better understand how PILOT uses privileged information, we investigate three questions: \textbf{(RQ1)} whether blind synthesis and label-prediction verification improve the quality of reasoning supervision; \textbf{(RQ2)} whether the posterior improves exploration of useful reasoning trajectories; and \textbf{(RQ3)} whether posterior discoveries transfer to the inference-time prior.
We conduct these analyses on 1,000 examples sampled approximately equally across all MMEB-V2 subsets, each comprising a query and its ground-truth positive candidate.
More details can be seen in Appendix~\ref{app:probe_dataset}.

\paragraph{RQ1. }
\begin{figure}[t]
\begin{center}
\includegraphics[width=1.0\textwidth]{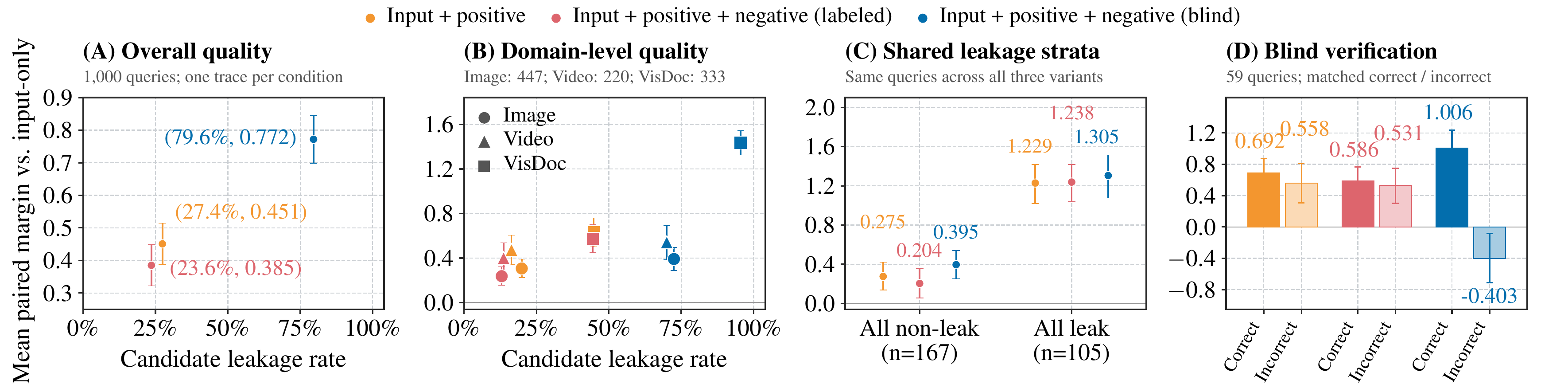}
\end{center}
\caption{\textbf{Reasoning quality, candidate leakage, and verification under different synthesis prompts.} (A–B) Overall and domain-level judged retrieval usefulness versus candidate leakage. (C) Comparison on shared query subsets with no leakage flagged under any variant or leakage flagged under every variant. (D) Comparison conditioned on blind-prediction correctness, with matched sample indices across conditions. }
\label{fig:anlysis_q1}
\end{figure}

We use the annotator to generate query-side reasoning under four conditions: input only, input with a positive candidate, input with a labeled positive–negative pair, and input with a blind positive–negative pair (our blind contrastive posterior). 
GPT-5.6 sol is used to assign retrieval-usefulness margins in $[-2,2]$ (higher is better) and separately flag candidate information unsupported by the query (Appendix~\ref{app:detailed_rq1}).
Figure~\ref{fig:anlysis_q1}(A--B) shows that blind synthesis achieves the highest mean margin ($0.772$), particularly on visual documents, but also the highest leakage rate ($79.6\%$).
It retains its advantage on the shared non-leakage subset (C), scoring $0.395$ versus $0.275$ for positive-only and $0.204$ for labeled-pair prompting.
Matched comparisons on 59 queries (D) yield margins of $1.006$ for correct blind predictions versus $-0.403$ for incorrect ones, with much smaller differences in the controls.
Blind synthesis thus elicits useful reasoning, while verification provides a quality filter.
Neither guarantees that the resulting traces are reproducible from the input alone, motivating transfer that also accounts for prior reachability.

\paragraph{RQ2. }

\begin{figure}[t]
\begin{center}
\includegraphics[width=1.0\textwidth]{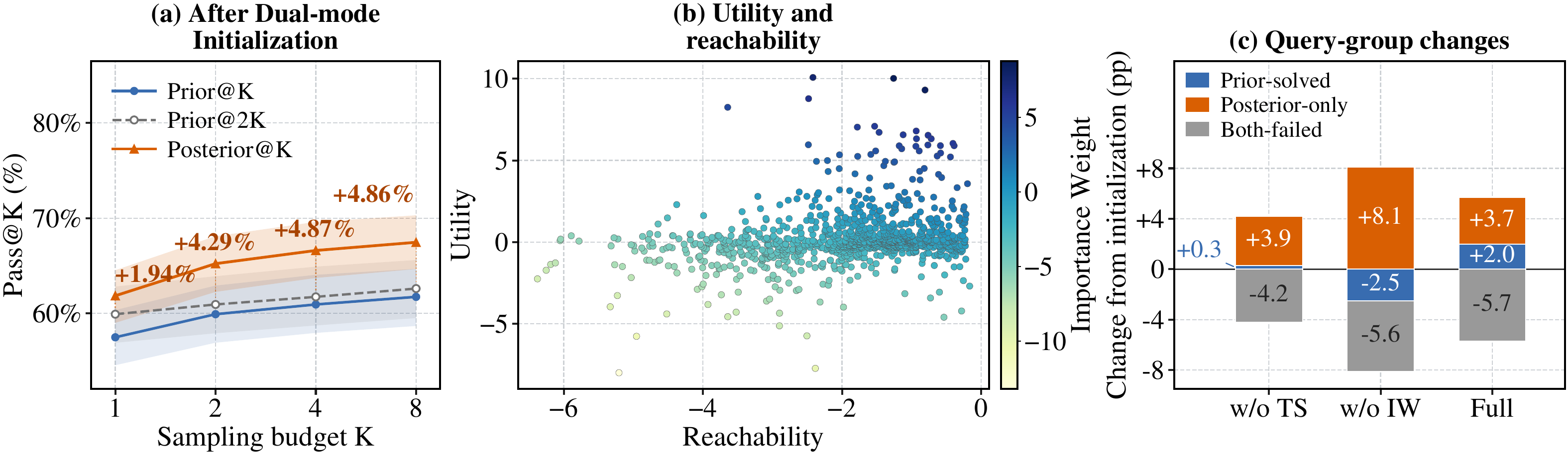}
\end{center}
\caption{Posterior exploration and transfer. (a) Pass@K after dual-mode initialization. (b) Utility–reachability trade-off, colored by transfer score. (c) Changes in prior-solved, posterior-only, and both-failed query proportions relative to dual-mode initialization comparing the full method and ablations without transfer supervision (w/o TS) or importance weighting (w/o IW).}
\label{fig:anlysis_q2&q3}
\end{figure}

We assess posterior exploration at the dual-mode SFT checkpoint by sampling 16 traces per query under each prior and blind posterior prompt, then replay all traces under the prior prompt and evaluate their embeddings
against the same candidate pool.
Figure~\ref{fig:anlysis_q2&q3} (a) shows that Posterior@$K$ outperforms both Prior@$K$ and Prior@$2K$ across all tested budgets.
As $K$ increases from 1 to 8, posterior Pass@$K$ rises from 61.8\% to 67.5\%, exceeding Prior@$2K$ by 1.9--4.9 percentage points.
Thus, conditioning exploration on relational evidence yields gains that doubling input-only sampling does not recover, motivating its use to guide reasoning exploration during training.

\paragraph{RQ3. }
We investigate this question from two complementary perspectives: how transfer prioritizes posterior trajectories and how transfer supervision and importance weighting affect retrieval coverage.
Figure~\ref{fig:anlysis_q2&q3}(b) plots one valid posterior trajectory per query by its utility and reachability components, colored by the un-normalized importance weight.
The distribution illustrates the trade-off encoded by our weighting: the retrieval utility increases the transfer priority, whereas a large reachability penalty can offset this benefit. Thus, high-reward posterior trajectories are not automatically assigned high transfer scores.
Panel (c) compares Full \ourmodel~with variants removing transfer supervision (w/o TS) or replacing importance weighting with uniform weighting (w/o IW).
Using eight traces per query and mode, we count success if any trace retrieves a relevant candidate at rank 1 and partition queries into prior-solved, posterior-only, and both-failed groups.
Compared with w/o IW, Full raises prior-solved coverage from 59.3\% to 63.8\% and lowers posterior-only coverage from 15.8\% to 11.4\%, with nearly unchanged coverage by either mode (75.1\% vs 75.2\%). Against w/o TS, Full improves prior-solved coverage by 1.7 points and reduces both-failed queries by 1.5 points. These aggregate results suggest that selective transfer helps make retrieval success accessible to the inference-time prior, whereas greater posterior-only coverage need not improve input-only retrieval.
More details of the experiment can be seen in Appendix~\ref{app:rq3}.

\section{Conclusion}
We presented \ourmodel, a latent-variable framework that uses training-time relational evidence to improve reasoning for multimodal embeddings while retaining input-only inference.
It outperforms main baselines across all 12 MMEB-V2 task groups at both 2B and 7B scales.
Our findings suggest that retrieval supervision should inform which reasoning trajectories are explored, alongside evaluating their embedding utility.
Yet increased candidate leakage and limited gains from direct distillation highlight the need for selective transfer.
The benefits of posterior adaptation and importance-weighted prior fitting point to a practical training strategy that jointly improves exploration and transfers its discoveries according to retrieval utility and prior reachability.
For settings with relational evidence available only during training, this provides a basis for developing better exploration policies and transfer objectives without requiring candidate context at inference.

\section*{AI use statement}
In this work, we used generative AI tools to polish manuscript drafts and assist with experimental code implementation.
We have reviewed all AI-assisted work. Specifically, we manually check the generated manuscript paragraphs for potential semantic deviation or grammar errors. 
LLM-generated experimental code was verified by the author manually and tested to ensure the correctness before being used in the practical experiments.
We take responsibility for the final content of this work, including text, claims or artifacts produced with the aid of generative AI.

\section*{Reproducibility statement}
Our experiments use publicly available datasets and backbone models.
We describe the training data, model configurations, optimization settings, and evaluation protocols in the experimental section, with additional implementation details in Appendix~\ref{app:implementation_details}.
Synthesis prompts and the reasoning evaluation protocol are provided in Appendix~\ref{app:prompt_templates}.
Appendix~\ref{app:mixture_iwae_derivation} details the derivation from the importance-weighted lower bound to the three gradient components instantiated in prior--posterior co-evolution.
About the further analysis experiment, Appendix~\ref{app:probe_dataset} documents probe dataset construction.
More detailed experiments about the model-based evaluation of the synthesized posterior reasoning trace can be seen in Appendix~\ref{app:detailed_rq1}.
More experiments about the effect of the importance weight are presented in Appendix~\ref{app:rq3}.
All the quantitative experiment results are averaged over three random seeds.
We will release our training and evaluation code, configurations, and synthesized reasoning data to facilitate reproduction.

\section*{Acknowledgments}
This work was supported by the DIGIS Information Extraction Project (DFG, German Research Foundation) under Grant 437919684, by the Lower Saxony Ministry of Science and Culture and the Volkswagen Foundation, and by the Deutsche Forschungsgemeinschaft (DFG, German Research Foundation) under Grant 564661959. 
The authors gratefully acknowledge the German Federal Ministry of Research, Technology and Space and the German federal states (\url{http://www.nhr-verein.de/en/our-partners}) for providing computational resources through the National High-Performance Computing (NHR) joint funding program, as well as the KISSKI project (\url{https://kisski.gwdg.de/en/leistungen/}) for providing additional computational resources that supported this work.



\bibliography{iclr2027_conference}
\bibliographystyle{iclr2027_conference}

\appendix
\section{Appendix}
\subsection{Implementation Details}
\label{app:implementation_details}

\subsubsection{Training Settings}
\paragraph{Supervised fine-tuning.} During SFT initialization, we freeze the vision encoder while updating the visual merger and language model without LoRA. The training objective combines autoregressive losses on the query and positive-candidate responses with symmetric contrastive losses on the normalized representations at the generative and discriminative embedding tokens. Both contrastive losses use cross-device in-batch negatives and a temperature of 1/30, and all four loss terms are summed with unit coefficients. We train for 5,000 optimizer steps using AdamW with a learning rate of $5\times10^{-5}$, zero weight decay, a global batch size of 1024 via gradient accumulation and $\epsilon=10^{-6}$. We use cosine learning-rate decay with a 3\% warmup and clip the gradient norm at 0.5. The per-device batch size is two query–candidate pairs, with six gradient-accumulation steps. Training uses a maximum sequence length of 12,288 tokens, bfloat16 precision, gradient checkpointing, and DeepSpeed ZeRO-3. We set both the training and data seeds to 42 and save checkpoints every 500 steps. The fine-tuning of the 2B and 7B model is conducted on 64 NVIDIA A100 80 GB GPUs. 

\paragraph{Coevolution.} We train for one epoch, comprising 192 rollout iterations, with a global batch size of 64 and eight sampled responses per input under each prompt mode. Responses are generated at temperature 1.0 without top-k or nucleus truncation, with maximum prompt and response lengths of 7,168 and 1,024 tokens, respectively. We use AdamW with a constant learning rate of $10^{-6}$, momentum coefficients (0.9,0.999), weight decay 0.01, and gradient clipping at 1.0. GRPO uses within-group standardized advantages, a policy-ratio clipping threshold of 0.2, and one optimization epoch per rollout batch. The prior GRPO, posterior GRPO, and weighted SFT objectives are normalized separately and combined with coefficients 1.0, 1.0, and 0.3, respectively. For both prior and posterior branch, we set $\lambda_{\mathrm{fmt}}=0.02$. We apply KL regularization against the initialization checkpoint with coefficient 0.05 and no entropy bonus. 
Retrieval rewards use an InfoNCE temperature of 0.2. 
\begin{equation}
R_{\mathrm{ret}}(\mathbf{e}_\theta(x,z_k), \mathcal{R})
=
\log
\frac{
  \sum_{m=1}^{M}
  \exp\!\left(
      \operatorname{sim}(\mathbf{e}_\theta(x,z_k),\mathbf{e}_\theta(x^+,z_m))/T
  \right)
}{
  \sum_{x^-\in \mathcal{X}^-}\sum_{m=1}^{M}
  \exp\!\left(
      \operatorname{sim}(\mathbf{e}_\theta(x,z_k),\mathbf{e}_\theta(x^-,z_m))/T
  \right)
},
\label{eq:retrieval_reward}
\end{equation}
Here and $M$ is the number of rollouts per candidate.
To induce the conditional likelihood, we set $\beta=-10$.
The similarity function \(\operatorname{sim}(\mathbf{e}_\theta(x,z_k),\mathbf{e}_\theta(x^+,z_m))=\mathbf{e}_\theta(x,z_k)^{\top}\mathbf{e}_\theta(x^+,z_m)\) computes cosine similarity, with temperature $T=0.2$. The reward is the log-probability assigned to the positive set; higher values indicate better retrieval performance.
Training runs on 16 NVIDIA A100 80GB GPUs for the 2B backbone and 32 NVIDIA A100 80 GB GPU for the 7B backbone using FSDP2, bfloat16 mixed precision, gradient checkpointing, and vLLM-based sampling.

\subsubsection{Evaluation Settings}
We evaluate on MMEB-V2 using the subset-specific task configurations and candidate pools provided by the its official repo. Both queries and candidates are independently encoded in generative mode using only their original inputs and task instructions, without additional positive or negative examples. For each input, we generate a single reasoning trace using greedy decoding (temperature 0) with a maximum of 1,024 new tokens. Our customized vLLM backend directly returns the $L_2$-normalized final-layer hidden state at the {<gen\_emb>} token as the embedding. Candidates are ranked by cosine similarity against either the complete subset-level candidate pool or the query-specific candidate set, as specified by the task configuration. We use an evaluation batch size of 96, image-resizing bounds of 784–2,359,296 pixels, and eight sampled frames for video inputs according to the task configuration. Retrieval metrics are computed separately for each subset using its ground-truth relevance annotations.

\subsubsection{Posterior Reasoning Trace Synthesis}
Posterior reasoning synthesis. We use GLM-4.1V-9B-Thinking as the annotation model, served through vLLM with a maximum context length of 32,768 tokens. We sample annotations with temperature 1.0, top-p 0.95, top-k 20, and a presence penalty of 1.5. For each query, the annotator receives its positive candidate and a randomly sampled negative candidate from the same source subset; symmetrically, each candidate is accompanied by a positive and a negative query. Auxiliary items are presented in randomized order without revealing their relevance labels. The annotator identifies the positive item and generates a content-grounded reasoning trace and concise summary, with instructions to avoid explicit option references and details unsupported by the focal input. We retain a posterior training pair only when positive-item identification is correct on both sides and the outputs pass format validation. Predicted option labels are excluded from the supervised responses, while reasoning and summaries are serialized into the training format ending with <gen\_emb>. The final SFT examples use a simplified, role-independent posterior prompt rather than the detailed annotation instructions.
\begin{table}[htp]
  \centering
  \caption{Composition of the dual-mode SFT dataset.
  Each pair contains a query and its positive candidate,
  with reasoning supervision on both sides.}
  \label{tab:sft_data_composition}
  \begin{tabular}{lrr}
      \toprule
      Mode & Number of pairs & Proportion \\
      \midrule
      Prior     & 787,202   & 53.79\% \\
      Posterior & 676,158   & 46.21\% \\
      \midrule
      Total     & 1,463,360 & 100.00\% \\
      \bottomrule
  \end{tabular}
\end{table}
Table~\ref{tab:sft_data_composition} summarizes the dataset used for dual-mode initialization. It contains 1,463,360 query–candidate pairs, comprising 787,202 prior-mode pairs (53.79\%) and 676,158 posterior-mode pairs (46.21\%). Each pair provides reasoning supervision for both the query and its positive candidate. Prior-mode prompts contain only the focal input and instructions, whereas posterior-mode prompts additionally include positive and negative auxiliary examples.

\subsection{Prompt Templates}
\label{app:prompt_templates}

We distinguish prompts consumed by the embedding model during SFT or RL from prompts sent to the expert annotation model. The former are intentionally short, whereas the latter contain detailed grounding and leakage-control instructions.
In the templates below, angle-bracketed uppercase strings such as \texttt{<MAIN\_TEXT>} and \texttt{<MAIN\_VISUAL>} are exposition-only placeholders. Literal model tokens, including \texttt{<disc\_emb>},
\texttt{<think>}, \texttt{<answer>}, and \texttt{<gen\_emb>}, are shown without substitution. Image and video objects are inserted at the positions marked by the corresponding visual placeholders.

\subsubsection{Embedding-model training prompts}

The prior receives only the focal input. The same template is used on the query and candidate sides, with the appropriate text and visual content substituted for the focal input.

\begin{promptbox}{Prior prompt used for embedding-model training}
<MAIN_VISUAL>
<MAIN_TEXT>
<disc_emb>
Represent the above input text, images, videos, or any combination of the three as embeddings.
First output the thinking process in <think> </think> tags and then summarize the entire input in a word or sentence. Finally, use the <gen_emb> tag to represent the entire input.
\end{promptbox}

The posterior receives the same focal input followed by one positive and one negative in randomized order. The labels do not expose which item is positive. For query-side training, the auxiliary items are candidate inputs; for candidate-side training, they are queries. Multimodal objects are interleaved at
the indicated positions.

\begin{promptbox}{Blind posterior prompt used for embedding-model training}
<MAIN_VISUAL>
<MAIN_TEXT>
<disc_emb>
Represent the above input text, images, videos, or any combination of the three as embeddings. A few auxiliary items are provided below in randomized order (each may also include text, images, or videos) -- one is a true match for the above input and the other is not. Infer which is which during your reasoning to identify the discriminative features. The embedding must represent the above input alone.

Auxiliary item [A]:
<AUXILIARY_A_VISUAL>
<AUXILIARY_A_TEXT>

Auxiliary item [B]:
<AUXILIARY_B_VISUAL>
<AUXILIARY_B_TEXT>

First output the thinking process in <think> </think> tags and then summarize the entire input in a word or sentence. Finally, use the <gen_emb> tag to represent the entire input.
Do not copy exact answer spans or unique details from the auxiliary items. Do not treat the auxiliary items as the object to be represented.
\end{promptbox}

\subsubsection{Expert prompts for SFT annotation}
The query-side blind posterior annotation prompt used for the multimodal SFT data is formed by concatenating the following input wrapper with
the query-side instruction. Candidate labels are randomized before the request is sent to the expert. Samples for which the returned choice is incorrect are discarded.

\begin{promptbox}{Expert blind posterior annotation wrapper: query side}
You are generating training data for a retrieval embedding model. You will be shown a QUERY and a few candidate documents labeled with [A] [B] [C] ... sequentially. One is a true positive match for the query; the other is not. Infer which is the true positive and write reasoning and an answer ABOUT THE QUERY.
The "choice" field is used to filter training data. "reasoning" and "answer" are used as training targets for the student model.

MULTIMODAL INPUT
Each side (query or candidate) may carry an image or a video, interleaved at its position in this message. The text and visual together constitute one side.
- When the text reads "describe the content of the image" / "describe the content of the video" or a similar placeholder, treat the accompanying visual as the primary content of that side.
- When the text is substantive, the visual -- if present -- is supplementary context.
- When no visual is present, the text is the entire content.

INPUT
Query:
<QUERY_TEXT_AND_VISUAL>

Candidate [A]:
<CANDIDATE_A_TEXT_AND_VISUAL>

Candidate [B]:
<CANDIDATE_B_TEXT_AND_VISUAL>
\end{promptbox}

\begin{promptbox}{Expert blind posterior annotation instruction: query side}
CHOICE FIELD
- "choice" identifies the candidate that is the true positive match for the query.
- Output exactly "A" or "B" or "C" or ... -- single letter, uppercase, no other text. Commit to one; do not output "neither" or "uncertain."
- A true positive satisfies the query's information need. If both seem plausible, pick the one with stronger and more specific alignment.
- Choice must be consistent with your internal reasoning, but reasoning/answer must NOT reveal which candidate you saw.

REASONING REQUIREMENTS
- Be content-rich: derive answers, compute values, identify entities, describe visible content concretely. Richer reasoning yields a stronger embedding.
- If the query has a derivable answer (factual, numerical, descriptive, identification), perform the derivation and state the result.
- The discriminative features that determined your choice should appear as positive specifications of what a true match must have -- NOT as comparisons between candidates.
- Overlap with either candidate's content is acceptable when independently derivable from the query alone. The leakage constraint applies only to content unreachable without seeing the candidates.
- Summarize your reasoning in the answer field with one or two sentence.

LEAKAGE -- your "reasoning" and "answer" must not:
- include information knowable only from seeing either candidate (textual or visual);
- describe, paraphrase, quote, or mirror either candidate's wording or visual specifics;
- use comparative phrasing ("the better match," "of the two," "rather than," "as opposed to") or reference multiplicity ("among the options," "either of them," "between them");
- name the focal's role or the auxiliary's role ("the query," "the user," "the candidate," "the match");
- introduce visual vocabulary when the query itself has no visual content;
- describe the query's own visual at greater specificity than it observably supports.

A reader given only "reasoning" and "answer" should be unable to tell whether you saw zero, one, or many candidates, and unable to infer visual features of any candidate.

STYLE
STYLE (for "reasoning" and "answer")
- Vary your opening sentence across examples.

OUTPUT
Return a single JSON object, nothing else:
{
  "choice": "A" or "B" or ...,
  "reasoning": "...",
  "answer": "..."
}
\end{promptbox}

\subsubsection{Updated candidate-side expert annotation prompts}
\label{app:candidate_annotation_v3}

The distinction between the query and candidate side is substantive: a text query should not acquire an unseen page's answer, whereas a candidate page should preserve its own visible names, facts, numbers, and evidence relationships. Auxiliary queries may redirect inspection, but must not supply absent content or reduce a general candidate representation to the answer to one particular query.

\begin{promptbox}{System prompt for candidate annotation}
You generate supervised reasoning targets for a candidate-side multimodal retrieval embedding model. Produce an informative semantic representation of the CANDIDATE, preserving its content and important details.

Distinguish candidate evidence from auxiliary query content. Queries may suggest interpretations or details worth checking; retain those suggestions only when the candidate independently supports them. Query selection is an intermediate verification task, and its label does not establish facts about the candidate.

Be concrete and faithful. Preserve useful information even when it is unnecessary for distinguishing the supplied queries. Describe observable content and evidence relationships instead of generic statements about retrieval or relevance. The target represents the candidate, not a response tailored to an auxiliary query.
\end{promptbox}

\begin{promptbox}{User-message wrapper for candidate annotation}
Generate a retrieval-oriented reasoning trace and embedding summary for the CANDIDATE.

MULTIMODAL INPUT
The candidate may contain text, an image, a video, or a combination. Its text and visual content jointly define the candidate.
- When the candidate text is a placeholder instruction, inspect the accompanying visual as the primary content.
- When the candidate text is substantive, preserve its entities, relations, modifiers, and constraints. Use any candidate visual as additional evidence.
- When the candidate is text-only, represent its stated content without importing an auxiliary query's image, video, reference object, or requested transformation.
- When the candidate is a document page, inspect the supplied page itself. Distinguish visible evidence from content that might exist on an unseen page.

CANDIDATE-FIRST ANALYSIS
Before evaluating any auxiliary query, identify the candidate's main entities, attributes, actions, relations, subject matter, and explicit claims. Inspect all available candidate modalities. Keep track of what is directly supported, what is a supported interpretation, and what remains unknown.

INPUT
Candidate:
<CANDIDATE_TEXT_AND_VISUAL>

<OPTIONAL_AUXILIARY_QUERY_BLOCK>

END OF INPUT

ANNOTATION INSTRUCTIONS

<CONDITION_WORKFLOW>

<SHARED_CANDIDATE_TARGET_RULES>

<OUTPUT_SCHEMA>
\end{promptbox}

\begin{promptbox}{Auxiliary-query block for blind candidate annotation}
Query [A]:
<QUERY_A_TEXT_AND_VISUAL>

Query [B]:
<QUERY_B_TEXT_AND_VISUAL>
\end{promptbox}

In the blind posterior, one query is positive for the focal candidate and the other is negative. Their order is randomized and relevance labels are not shown. The prompt does not assert that the negative is hard: the same wording supports either random or mined negatives. The returned choice identifies the positive query; correctness filtering, when applied, is a separate processing step.

\begin{promptbox}{Candidate blind posterior annotation workflow}
BLIND VERIFICATION AND CANDIDATE SYNTHESIS
The goal is an informative semantic representation of the candidate. Query selection is an intermediate verification task. The candidate is a true positive for one of Query [A] and Query [B].

1. Understand the candidate before evaluating either query.
Identify its entities, attributes, actions, relations, explicit claims, and main subject matter. Inspect relevant candidate visuals. Preserve both what is supported and what remains unknown.

2. Evaluate A and B independently.
For each query, check which requirements the candidate satisfies, which it contradicts, and which cannot be verified from the supplied content. Prioritize explicit entity, relation, modifier, and temporal constraints over broad topical similarity. An unknown match is not confirmed support; an assumption in a query is not evidence about the candidate.
Respect the retrieval unit: a document cover can establish document identity without displaying the requested internal fact. Missing evidence in a preview is not automatically a contradiction, but it does not license claims about unseen content.

3. Investigate meaningful ambiguities.
When the queries suggest different interpretations, return to the candidate and inspect evidence that could resolve the difference. Use this comparison to discover overlooked candidate details. If a difference is already obvious or irrelevant to the candidate's content, do not force it into the final representation.

4. Make the best A/B choice.
Choose the query whose information need the candidate satisfies more specifically and completely. Do not revise the candidate's content to accommodate a query. If the available evidence cannot resolve the choice reliably, still return the best choice, but leave the unresolved interpretation out of the final reasoning. A choice does not make uncertain query assumptions true.

5. Synthesize the candidate representation.
Return to the candidate content identified before the comparison. Integrate query-suggested details only when independently supported by the candidate. Preserve useful connections between entities, attributes, actions, claims, and evidence.
The representation should remain informative for many plausible retrieval queries, not merely sufficient to choose between A and B. Do not include the query-selection process in reasoning or answer. Do not replace a broad candidate summary with a short answer to the selected query.
\end{promptbox}

\begin{promptbox}{Candidate-grounding and output rules}
CANDIDATE-GROUNDED TARGET
1. Preserve the full retrieval-relevant meaning of the candidate. Retain its important content even when some details are unnecessary for a local query decision.
2. Follow the condition-specific workflow internally; do not narrate the annotation workflow in reasoning or answer.
3. In reasoning, describe concrete supported entities, attributes, actions, relations, claims, and spatial or temporal cues. For documents, preserve visible evidence relationships, including the association of a value with its entity, metric, unit, or time period when readable.
4. Include useful discriminative details when supported, without forcing a fixed number of details or inventing specificity. The representation should remain informative for many plausible retrieval queries.
5. The answer is a concise semantic embedding summary of the candidate. Preserve its central content and important qualifiers; do not merely repeat the task instruction or answer an auxiliary query.
6. Preserve uncertainty locally: omit or qualify an uncertain detail while keeping other supported content specific and informative.

USEFUL OUTPUT CONTENT
Retain:
- Concrete details directly supported by the candidate text or visual, including relevant names, numbers, dates, and factual answers actually present in the candidate.
- Interpretations supported by identifiable evidence within the candidate. Connect an interpretation to the evidence supporting it.
- Meaningful relationships between depicted objects, actions, events, or stated facts.
- For document pages, readable headings, entities, claims, units, tables, charts, and their relationships when informative. Describe layout only when it helps identify or interpret content; do not replace semantic content with generic layout descriptions.
- For videos, observable action states, event order, and temporal relationships supported by the available frames. Do not invent motion or events between unobserved frames.

Exclude:
- Query-specific names, numbers, dates, locations, requested changes, or assumptions not independently established by the candidate.
- Claims about unseen pages, obscured text, hidden image regions, unobserved events, or unavailable audio.
- Broad task descriptions that replace concrete candidate content, such as 'retrieve relevant information' or 'focus on visual features'.
- Fine-grained visual, OCR, spatial, numerical, or temporal details that cannot be inspected reliably.

A question's premise does not become a candidate fact. Conversely, a factual answer visible in the candidate should not be suppressed merely because an auxiliary query asks for it. Preserve the fact in its candidate-supported context, not as a narrated response to the query. Do not compute a new value solely to satisfy an auxiliary request unless the calculation and all its inputs are independently justified by the candidate; clearly distinguish a supported derivation from an explicitly stated value.

FINAL CHECK
- Remove references to auxiliary queries, A/B labels, selection, rejection, and the annotation procedure from reasoning and answer. A separately required top-level choice field is exempt.
- Preserve comparisons, negations, exclusions, and relationships that are explicitly part of the candidate itself.
- For every factual clause, identify its support in the candidate. Remove clauses supported only by auxiliary content.
- Check that no important candidate entity, action, relation, claim, or qualifier disappeared while following the condition-specific workflow.
- The final reasoning and answer must stand on their own as an analysis of the candidate after every auxiliary query is removed.
\end{promptbox}

\begin{promptbox}{Candidate blind posterior output schema}
OUTPUT
Return one JSON object and nothing else. Set choice to exactly "A" or "B", a single uppercase letter identifying the positive query:

{
  "choice": "A",
  "reasoning": "...",
  "answer": "..."
}
\end{promptbox}

\subsubsection{Diagnostic annotation prompt-variant experiment}
\label{app:diagnostic_prompts}

The four conditions are input only, input + positive, input + positive + negative (labeled), and input + positive + negative (blind). All four include the query input.
The negative is sampled from the same subset rather than selected by hard-negative mining. In the blind condition, the positive and negative are assigned randomized A/B labels without revealing their ground-truth relevance.

All conditions share the system prompt, query-first introduction, task adapter, and target rules below. Only the auxiliary-input blocks, condition-specific workflow, and requirement for a separate choice field differ. 
The text is transcribed from the generation code, not rewritten for presentation.
The uppercase angle-bracketed placeholders indicate inserted content; bracketed condition headings in the candidate-block box are explanatory and are not sent to the expert. 
The query and candidates can each contain text and visual inputs.
The explicit \texttt{END OF INPUT} boundary separates the final input from
\texttt{ANNOTATION INSTRUCTIONS}, preventing candidate text from running into
the task-adapter heading.

\begin{promptbox}{Candidate blocks for the four diagnostic conditions}
[INPUT ONLY]
No candidate block is included.

[INPUT + POSITIVE]
Positive candidate:
<POSITIVE_TEXT_AND_VISUAL>

[INPUT + POSITIVE + NEGATIVE (LABELED)]
Positive candidate:
<POSITIVE_TEXT_AND_VISUAL>

Negative candidate:
<NEGATIVE_TEXT_AND_VISUAL>

[INPUT + POSITIVE + NEGATIVE (BLIND)]
Candidate [A]:
<CANDIDATE_A_TEXT_AND_VISUAL>

Candidate [B]:
<CANDIDATE_B_TEXT_AND_VISUAL>
\end{promptbox}

\begin{promptbox}{Diagnostic workflow: Input only}
DIRECT QUERY WORKFLOW
Analyze the query directly. Cover its most retrieval-relevant observable content, constraints, and evidence relationships. Preserve what is known and what is requested without inventing missing facts.
\end{promptbox}

\begin{promptbox}{Diagnostic workflow: Input + positive}
POSITIVE-GUIDED WORKFLOW
First establish the query's meaning and important details independently.
Use the positive candidate to suggest interpretations, overlooked details, or evidence relationships worth checking. Return to the query to verify each suggestion. The positive label establishes a retrieval match, not independent support for every candidate fact.
Integrate verified suggestions with the query's other important content. If none can be verified, use a detailed direct analysis of the query. Do not narrow the query to the positive candidate's incidental content.
\end{promptbox}

\begin{promptbox}{Diagnostic workflow: Input + positive + negative (labeled)}
LABELED CONTRASTIVE WORKFLOW
First establish the query's meaning and important details independently.
Use the supplied positive and negative labels. Inspect their differences to identify meaningful ambiguities or overlooked query evidence. Return to the query to verify any proposed distinction.
If the negative differs only in broad topic, or its differences add nothing useful to understanding the query, do not force a contrast into the final representation. Continue with positive-guided inspection and direct query analysis.
Retain verified discriminative properties alongside the query's other important entities, actions, relations, and constraints. Do not reduce the representation to the minimum features needed to distinguish these two candidates. Express retained information as self-contained query content or requirements.
\end{promptbox}

\begin{promptbox}{Diagnostic workflow: Input + positive + negative (blind)}
BLIND VERIFICATION AND QUERY SYNTHESIS
The goal is an informative semantic representation of the query. Candidate selection is an intermediate verification task.

1. Understand the query before evaluating either candidate.
Identify its entities, attributes, actions, relations, and explicit constraints. Inspect relevant query visuals. Preserve both what is specified and what remains unknown.

2. Evaluate A and B independently.
For each candidate, check which query requirements it satisfies, which it contradicts, and which cannot be verified from the supplied content. Prioritize explicit entity, relation, modifier, and temporal constraints over broad topical similarity. An unknown match is not confirmed support; extra candidate detail is not evidence of correctness.
Respect the retrieval unit: a document cover can establish document identity without displaying the requested internal fact. Missing evidence in a preview is not automatically a contradiction.

3. Investigate meaningful ambiguities.
When the candidates suggest different interpretations, return to the query and inspect evidence that could resolve the difference. Use this comparison to discover overlooked query details. If a difference is already obvious or irrelevant to the query's meaning, do not force it into the final representation.

4. Make the best A/B choice.
Do not revise the query's meaning to accommodate a candidate. If the available evidence cannot resolve the choice reliably, still return the best choice, but leave the unresolved interpretation out of the final reasoning. A choice does not make uncertain candidate details true.

5. Synthesize the query representation.
Return to the query content identified before the comparison. Integrate candidate-suggested details only when independently supported by the query. Preserve useful connections between entities, attributes, actions, and constraints.
The representation should remain informative among many plausible retrieval candidates, not merely sufficient to choose between A and B. Do not include the candidate-selection process in reasoning or answer.
\end{promptbox}

\begin{promptbox}{Shared query-grounding and output rules for diagnostic generation}
QUERY-GROUNDED TARGET
1. Preserve the full retrieval-relevant meaning of the query. Retain its important content even when some details are unnecessary for a local candidate decision.
2. Follow the condition-specific workflow internally; do not narrate the annotation workflow in reasoning or answer.
3. In reasoning, describe concrete supported entities, attributes, actions, relations, and spatial or temporal cues. For information-seeking queries, explain the evidence relationships needed to resolve the request.
4. Include useful discriminative details when supported, without forcing a fixed number of details or inventing specificity. The representation should remain informative among many plausible retrieval candidates.
5. The answer is a concise semantic embedding summary. Preserve the central meaning and constraints; do not merely repeat the task instruction.
6. Preserve uncertainty locally: omit or qualify an uncertain detail while keeping other supported content specific and informative.

USEFUL OUTPUT CONTENT
Retain:
- Concrete details directly supported by the query text or visual.
- Interpretations supported by identifiable evidence within the query. Connect an interpretation to the evidence supporting it.
- Evidence relationships implied by the request, such as linking an entity to an attribute, comparing specified years, or locating the numerator and denominator of a requested ratio.

Exclude:
- Candidate-specific answers, names, numbers, quotations, dates, or locations that cannot be established from the query.
- Claims that a particular chart, table, page, or layout must exist solely because it appears in a candidate. A plausible evidence format may be described as a possibility, not an observed fact.
- Broad task descriptions that replace concrete query content, such as 'retrieve relevant information' or 'focus on visual features'.
- Fine-grained visual, OCR, spatial, or temporal details that cannot be inspected reliably.

For information-seeking queries, preserve the requested fact type and relationship. State a factual answer only when the query itself supports it. A candidate's answer does not become a query fact.

FINAL CHECK
- Remove references to auxiliary candidates, A/B labels, selection, rejection, and the annotation procedure from reasoning and answer. A separately required top-level choice field is exempt.
- Preserve comparisons, negations, exclusions, and transformations that are explicitly part of the query itself.
- For every factual clause, identify its support in the query. Remove clauses supported only by auxiliary content.
- Check that no important query entity, action, relation, or constraint disappeared while following the condition-specific workflow.
- The final reasoning and answer must stand on their own as an analysis of the query.
\end{promptbox}

\begin{promptbox}{Non-blind output schema}
OUTPUT
Return one JSON object and nothing else:

{
  "reasoning": "...",
  "answer": "..."
}
\end{promptbox}

\begin{promptbox}{Blind output schema}
OUTPUT
Return one JSON object and nothing else:

{
  "choice": "A",
  "reasoning": "...",
  "answer": "..."
}
\end{promptbox}

The task adapter is selected from the following exact strings according to the
query's task type. Only one adapter is inserted into each request. The general
retrieval adapter is the fallback when no task-specific adapter applies.

\begin{promptbox}{Task adapters for diagnostic generation}
TASK ADAPTER: CLASSIFICATION
Describe concrete query-visible category evidence such as shape, parts, material, texture, scene context, behavior, or composition. Do not stop at saying that the input needs classification. A category hypothesis is useful only when accompanied by visible query evidence.

TASK ADAPTER: QUESTION ANSWERING
Preserve exactly what is being asked and identify the query evidence needed to answer it. Inspect the query visual for relevant entities, attributes, relations, text, or events. State a proposed answer only if the query itself supports it; otherwise represent the requested answer type and evidence without guessing.

TASK ADAPTER: TEXT-TO-IMAGE RETRIEVAL
Preserve named entities, objects, actions, scene, style, modifiers, and relations from the text. Describe the concrete visual evidence a relevant image should satisfy without inventing details absent from the query.

TASK ADAPTER: IMAGE-TO-TEXT RETRIEVAL
Describe concrete observable image content and the textual semantics that should represent it, including salient entities, actions, scene, and relations.

TASK ADAPTER: COMPOSED IMAGE RETRIEVAL
Separate what must be retained from the reference image from what the modification text changes. State both retained visual identity and requested transformation concretely.

TASK ADAPTER: VISUAL GROUNDING
Represent the referred entity using query-visible attributes and spatial relations. Include location or relation cues that distinguish the referent from nearby alternatives.

TASK ADAPTER: TEXT-TO-VIDEO RETRIEVAL
Preserve entities, actions, event order, setting, and temporal qualifiers. Describe the observable event a relevant video should contain.

TASK ADAPTER: VIDEO MOMENT RETRIEVAL
Represent the target event, participating entities, action state, temporal order, and useful start/end cues supported by the query.

TASK ADAPTER: VISUAL DOCUMENT RETRIEVAL
Preserve the query's exact entities, qualifiers, requested fact type, and evidence relationship. Describe what a relevant page must establish without inventing the page's answer, OCR text, layout, table values, or chart values.

TASK ADAPTER: GENERAL RETRIEVAL
Preserve concrete query content, relations, constraints, and the evidence a relevant result should satisfy.
\end{promptbox}

\subsubsection{Blinded trace-audit instruction}
\label{app:trace_audit_instruction}

This is a model-based qualitative judgment.
The boxes below reproduce the evidence, rating, and per-trace flag instructions from the operational review protocol. Each packet provides the full query, positive, random negative, available media, and anonymous left/right traces.
Variant names and predicted choice labels are masked, although the prose can sometimes reveal the condition. The judge model considers reasoning and answer together. 
The signed margin measures judged query-grounded retrieval usefulness, not an observed change in retrieval accuracy.

The expanded audit also records that prior results had been unblinded before expansion; it should therefore not be described as fully independent blinding.
Leakage flags include explicit narration of candidate selection or comparison, not only the importation of candidate-only facts. These flags inform but do not mechanically determine the pair rating.

\begin{promptbox}{Operational audit instruction: evidence and masking}
Read every assigned source's full query, positive, random negative, reasoning,
and answer. Display and inspect its actual media, including sampled video
frames, not merely paths or encoded bytes. Enlarge/crop original document pages
when the contact sheet cannot support a claimed detail. A contact sheet with
unreadable OCR does not establish that an OCR claim is false. Record limitations.
Do not look at annotations.jsonl, private_pair_mapping.jsonl, old score files,
or occurrence_mapping.json while scoring. Variant names and predicted choice
labels are masked in public packets, though prose can sometimes reveal them.

Packets are in canonical/packets/source_NNNN.json. They contain distinct
trace IDs and left/right pairs; exact overlaps across the two audit designs
have been deduplicated. Never infer ratings from length, keywords, or variant
identity. Reading via a tool must return full text; reread any truncated output.
\end{promptbox}

\begin{promptbox}{Operational audit instruction: paired ratings}
Assess usefulness as query-grounded reasoning for retrieval after the auxiliary
candidates are removed. The observed corpus positive and negative help evaluate
relevance; do not reward simply guessing a candidate or repeating its label.

- L2/R2: the indicated side is clearly better, with an important grounded gain
  or correction of a central entity, action, relation, constraint, omission,
  factual error, or candidate-dependent answer. Explain the concrete difference.
- L1/R1: a smaller but meaningful grounded gain or reduction in misleading
  content, not stylistic preference alone.
- T0: no meaningful net difference. Equal quality can be equally good or bad.

Consider the reasoning and answer together. Correct contextual interpretation
can be useful even when the answer is concise. Longer is not automatically
better; extra faithful detail is not automatically harmful. A directly grounded
class name is not automatically leakage, and naming a class already well
described by the other trace is not automatically a win. Preserve meaningful
uncertainty in ambiguous queries. Do not count sample difficulty as a causal
effect of verification. Genuine query negations/comparisons are permitted.
\end{promptbox}

\begin{promptbox}{Operational audit instruction: per-trace flags}
Enter all three flags as explicit booleans once per distinct trace:

- candidate_only_leak: final reasoning/answer imports concrete auxiliary-only
  information not independently supported by the query, or explicitly narrates
  candidate identities, labels, choice, rejection, or auxiliary comparison.
  A queried fact, plausible evidence relationship, or visibly supported class
  hypothesis is not leakage merely because it aligns with the positive.
- generic_or_vacuous: the output is predominantly reusable task boilerplate
  and fails to retain this query's concrete content/requirements. A short but
  specific summary is not vacuous.
- unsupported_claim: a substantive factual clause is contradicted or not
  established by available evidence and is presented as fact; includes invented
  visual/OCR/action relations. Clearly qualified plausible interpretations are
  not automatically false. Flags can co-occur. Candidate-supported but
  query-unsupported claims can carry both leakage and unsupported flags when
  asserted as facts about the query.

Flags inform, but do not mechanically determine, the pair rating. Describe
tradeoffs explicitly. Identical traces retain identical flags across all pairs
and repeated analyses. Identical left/right text must receive T0, not blank flags.
\end{promptbox}

\subsection{Derivation of the Mixture-IWAE Gradient}
\label{app:mixture_iwae_derivation}

We derive the gradient of the mixture importance-weighted objective used in
\ourmodel. For conciseness, we suppress the conditioning variables and define
\begin{equation}
    p_\theta(z) \coloneqq p_\theta(z\mid x),
    \qquad
    q_\theta(z) \coloneqq q_\theta(z\mid x,u),
    \qquad
    m_\theta(z)
    \coloneqq
    \alpha p_\theta(z)+(1-\alpha)q_\theta(z),
\end{equation}
where the mixture coefficient $\alpha\in(0,1)$ is independent of $\theta$.
We further define
\begin{equation}
    w_k
    =
    \frac{
        p_\theta(z_k)\,p_R(y\mid x,z_k)
    }{
        m_\theta(z_k)
    },
    \qquad
    \widehat{Z}_K
    =
    \frac{1}{K}\sum_{k=1}^{K}w_k,
    \qquad
    \bar{w}_k
    =
    \frac{w_k}{\sum_{j=1}^{K}w_j}.
\end{equation}
Using the reward-induced likelihood
$p_R(y\mid x,z)\propto\exp(\beta R_{\mathrm{ret}}(x,z))$,
the mixture-IWAE objective is
\begin{equation}
    \mathcal{J}_{\mathrm{IW}}
    =
    \mathbb{E}_{z_{1:K}\sim m_\theta}
    \left[
        \log\widehat{Z}_K
    \right]
    \leq
    \log p_\theta(y\mid x).
    \label{eq:app_iwae}
\end{equation}
We assume that $R_{\mathrm{ret}}$ is treated as a stop-gradient reward.
Consequently, $p_R(y\mid x,z)$ has no direct gradient with respect to
$\theta$.

\paragraph{Step 1: Differentiate the sampling distribution.}
Because the samples are drawn from the parameterized mixture
$m_\theta$, the score-function identity gives
\begin{align}
    \nabla_\theta\mathcal{J}_{\mathrm{IW}}
    &=
    \mathbb{E}_{m_\theta^K}
    \left[
        \log\widehat{Z}_K
        \nabla_\theta
        \log\prod_{k=1}^{K}m_\theta(z_k)
        +
        \nabla_\theta\log\widehat{Z}_K
    \right]
    \nonumber\\
    &=
    \mathbb{E}_{m_\theta^K}
    \left[
        \sum_{k=1}^{K}
        \log\widehat{Z}_K
        \nabla_\theta\log m_\theta(z_k)
        +
        \nabla_\theta\log\widehat{Z}_K
    \right].
    \label{eq:app_score_gradient}
\end{align}

\paragraph{Step 2: Differentiate the importance-weighted estimator.}
The gradient of the log-average importance weight is
\begin{align}
    \nabla_\theta\log\widehat{Z}_K
    &=
    \frac{
        \sum_{k=1}^{K}\nabla_\theta w_k
    }{
        \sum_{j=1}^{K}w_j
    }
    =
    \sum_{k=1}^{K}
    \bar{w}_k
    \nabla_\theta\log w_k.
    \label{eq:app_log_iw_gradient}
\end{align}
Under the stop-gradient reward assumption,
\begin{equation}
\begin{aligned}
    \nabla_\theta\log w_k
    &=
    \nabla_\theta
    \left[
        \log p_\theta(z_k)
        +
        \log p_R(y\mid x,z_k)
        -
        \log m_\theta(z_k)
    \right]
    \\
    &=
    \nabla_\theta\log p_\theta(z_k)
    -
    \nabla_\theta\log m_\theta(z_k).
\end{aligned}
\label{eq:app_log_weight_gradient}
\end{equation}
Substituting Eq.~\ref{eq:app_log_weight_gradient} into
Eq.~\ref{eq:app_log_iw_gradient} gives
\begin{equation}
    \nabla_\theta\log\widehat{Z}_K
    =
    \sum_{k=1}^{K}
    \bar{w}_k\nabla_\theta\log p_\theta(z_k)
    -
    \sum_{k=1}^{K}
    \bar{w}_k\nabla_\theta\log m_\theta(z_k).
    \label{eq:app_iw_gradient_expanded}
\end{equation}

Combining Eqs.~\ref{eq:app_score_gradient} and
\ref{eq:app_iw_gradient_expanded}, we obtain
\begin{equation}
\begin{aligned}
    \nabla_\theta\mathcal{J}_{\mathrm{IW}}
    =
    \mathbb{E}_{m_\theta^K}
    \left[
        \sum_{k=1}^{K}
        \left(
            \log\widehat{Z}_K-\bar{w}_k
        \right)
        \nabla_\theta\log m_\theta(z_k)
        +
        \sum_{k=1}^{K}
        \bar{w}_k\nabla_\theta\log p_\theta(z_k)
    \right].
\end{aligned}
\label{eq:app_two_part_gradient}
\end{equation}

A baseline $b_k(z_{-k})$ that does not depend on $z_k$ may be subtracted
from the score-function coefficient because
\begin{equation}
    \mathbb{E}_{z_k\sim m_\theta}
    \left[
        b_k(z_{-k})
        \nabla_\theta\log m_\theta(z_k)
    \right]
    =0.
\end{equation}
We therefore define the mixture score-function signal as
\begin{equation}
    A_k^{\mathrm{mix}}
    \coloneqq
    \log\widehat{Z}_K
    -
    \bar{w}_k
    -
    b_k(z_{-k}).
    \label{eq:app_mixture_advantage}
\end{equation}
The baseline can, for example, be a leave-one-out estimate constructed
from the remaining $K-1$ trajectories.

\paragraph{Step 3: Decompose the mixture score.}
Define the posterior responsibilities of the two mixture components:
\begin{equation}
    \rho_k^{p}
    =
    \frac{
        \alpha p_\theta(z_k)
    }{
        m_\theta(z_k)
    },
    \qquad
    \rho_k^{q}
    =
    \frac{
        (1-\alpha)q_\theta(z_k)
    }{
        m_\theta(z_k)
    },
    \qquad
    \rho_k^{p}+\rho_k^{q}=1.
\end{equation}
The score of the mixture can then be decomposed as
\begin{align}
    \nabla_\theta\log m_\theta(z_k)
    &=
    \frac{
        \alpha\nabla_\theta p_\theta(z_k)
        +(1-\alpha)\nabla_\theta q_\theta(z_k)
    }{
        m_\theta(z_k)
    }
    \nonumber\\
    &=
    \rho_k^{p}\nabla_\theta\log p_\theta(z_k)
    +
    \rho_k^{q}\nabla_\theta\log q_\theta(z_k).
    \label{eq:app_mixture_score}
\end{align}
Substituting Eq.~\ref{eq:app_mixture_score} into
Eq.~\ref{eq:app_two_part_gradient} yields the exact three-term
decomposition
\begin{align}
    \nabla_\theta\mathcal{J}_{\mathrm{IW}}
    ={}&
    \underbrace{
    \mathbb{E}_{m_\theta^K}
    \left[
        \sum_{k=1}^{K}
        A_k^{\mathrm{mix}}\rho_k^{p}
        \nabla_\theta\log p_\theta(z_k)
    \right]
    }_{\text{prior policy improvement}}
    \nonumber\\
    &+
    \underbrace{
    \mathbb{E}_{m_\theta^K}
    \left[
        \sum_{k=1}^{K}
        A_k^{\mathrm{mix}}\rho_k^{q}
        \nabla_\theta\log q_\theta(z_k)
    \right]
    }_{\text{posterior policy improvement}}
    \nonumber\\
    &+
    \underbrace{
    \mathbb{E}_{m_\theta^K}
    \left[
        \sum_{k=1}^{K}
        \bar{w}_k
        \nabla_\theta\log p_\theta(z_k)
    \right]
    }_{\text{importance-weighted prior learning}}.
    \label{eq:app_three_part_gradient}
\end{align}

The first two terms arise because the prior and posterior jointly define
the parameterized proposal from which latent reasoning trajectories are
explored. The third term arises from the prior in the numerator of the
importance weight and increases the prior likelihood of trajectories
that receive high normalized importance weight.

\paragraph{Step 4: Recover the compact expression.}
For any function $f(z)$, the mixture responsibilities satisfy
\begin{equation}
    \mathbb{E}_{z\sim m_\theta}
    \left[
        \rho^{p}(z)f(z)
    \right]
    =
    \alpha
    \mathbb{E}_{z\sim p_\theta}[f(z)],
    \qquad
    \mathbb{E}_{z\sim m_\theta}
    \left[
        \rho^{q}(z)f(z)
    \right]
    =
    (1-\alpha)
    \mathbb{E}_{z\sim q_\theta}[f(z)].
    \label{eq:app_change_measure}
\end{equation}
Therefore, after absorbing $\alpha$, $1-\alpha$, the sum over $K$
samples, and the expectation over the remaining trajectories into the
definitions of $A^{p}$ and $A^{q}$, Eq.~\ref{eq:app_three_part_gradient}
can be written compactly as
\begin{equation}
    \nabla_\theta\mathcal{J}_{\mathrm{IW}}
    =
    \underbrace{
        \mathbb{E}_{p_\theta}
        \left[
            A^{p}\nabla_\theta\log p_\theta
        \right]
    }_{\text{prior improvement}}
    +
    \underbrace{
        \mathbb{E}_{q_\theta}
        \left[
            A^{q}\nabla_\theta\log q_\theta
        \right]
    }_{\text{posterior improvement}}
    +
    \underbrace{
        \mathbb{E}_{m_\theta}
        \left[
            \bar{w}\nabla_\theta\log p_\theta
        \right]
    }_{\text{weighted prior learning}}.
    \label{eq:app_compact_gradient}
\end{equation}
This is the compact gradient expression reported in the main text.

\paragraph{Connection to the practical objective.}
The first two terms are score-function policy gradients for the prior and posterior. In practice, we replace their mixture learning signal $A^{\mathrm{mix}}_k$ with group-relative advantages computed from the inference-compatible reward $R_k$ and optimize them with GRPO, which reduces variance.
This substitution is an approximation rather than an unbiased estimator of Eq. \ref{eq:app_three_part_gradient}. Since $A^{\mathrm{mix}}_k$ is computed from  $\log w_k = \log p_\theta(z_k) + \beta R_{\mathrm{ret}}(x, z_k) - \log m_{\theta}(z_k)$, it favors trajectories that are both useful and reachable under the prior, whereas the GRPO advantage retains only the reward. In the practical objective, reachability therefore enters only through the third term, which we implement as an importance-weighted likelihood loss,
\begin{equation}
    \mathcal{L}_{\mathrm{IW}}
    =
    -
    \mathbb{E}_{z_{1:K}\sim m_\theta}
    \left[
        \sum_{k=1}^{K}
        \operatorname{sg}[\bar{w}_k]
        \log p_\theta(z_k\mid x)
    \right],
\end{equation}
where $\operatorname{sg}[\cdot]$ stops gradients through the normalized
importance weights. The resulting practical objective combines prior
GRPO, posterior GRPO, and importance-weighted prior learning.

\paragraph{Remark on differentiable rewards.}
If gradients are propagated through the retrieval reward, then
\begin{equation}
    \nabla_\theta\log w_k
    =
    \nabla_\theta\log p_\theta(z_k)
    -
    \nabla_\theta\log m_\theta(z_k)
    +
    \beta\nabla_\theta R_{\mathrm{ret}}(x,z_k),
\end{equation}
and the exact gradient contains the additional term
\begin{equation}
    \beta\,
    \mathbb{E}_{m_\theta^K}
    \left[
        \sum_{k=1}^{K}
        \bar{w}_k
        \nabla_\theta R_{\mathrm{ret}}(x,z_k)
    \right].
\end{equation}
The three-term decomposition in
Eq.~\ref{eq:app_three_part_gradient} therefore assumes that the
retrieval reward is detached, as is standard in policy-gradient
optimization. If the alternative temperature parameterization
$p_R\propto\exp(R_{\mathrm{ret}}/\beta)$ is used, the derivation remains
unchanged after replacing $\beta$ by $1/\beta$.

\subsection{Probe Dataset Construction}
\label{app:probe_dataset}
We construct \textsc{MMEB-2-Probe}, a compact diagnostic dataset for controlled mechanism analyses. The probe contains 1,000 query--candidate triplets drawn from all 78 subsets of MMEB-2, with a nearly uniform allocation of 12 or 13 queries per subset. This prevents large subsets from dominating the analysis while preserving coverage of all benchmark domains and task types. The resulting dataset contains 447 image, 220 video, and 333 visual-document queries. The image portion covers classification, question answering, image-to-text retrieval, text-to-image retrieval, composed-image retrieval, and visual grounding; the video portion covers classification, question answering, text-to-video retrieval, and moment retrieval; and the visual-document portion covers document retrieval. Each example retains its source subset, task type, benchmark row index, and original text, image, or video modalities.

For each query, we construct a subset-specific candidate pool containing candidates that serve as official positives for other queries. Queries are then deterministically sampled within each subset according to the assigned quota. After excluding every official positive associated with the current query, we uniformly sample one candidate as its random negative. Sampling is deterministic and performed independently across queries, allowing the same candidate to be selected for multiple queries.
Among the 730 sampled negatives present in the saved retrieval rankings, the median rank is 316.5 and only 50 appear within the top 10, confirming that the resulting negatives are not systematically selected for difficulty.
The corresponding query, positive, and negative contents are loaded through the official MMEB-2 dataset loaders, ensuring that the stored multimodal inputs match those used during benchmark evaluation.

The final probe therefore provides one query, one official positive, and one qrel-safe random negative per instance. Retrieval rewards or other relative metrics are computed against the expanded subset-level candidate pool rather than against this single negative alone. Because the probe is sampled from MMEB-2 evaluation data, it is used exclusively for diagnostic and mechanism analyses and is never used for model training, hyperparameter selection, or checkpoint selection.

\subsection{Detailed RQ1 Experiment}
\label{app:detailed_rq1}
This experiment examines whether privileged candidate information helps an annotation model produce higher-quality query-side reasoning traces. In particular, we ask whether \textbf{a positive candidate}, \textbf{a labeled positive--negative pair}, or \textbf{an unlabeled candidate pair} can reveal retrieval-relevant information that is difficult to identify from the query alone. 
We additionally distinguish transferable improvements, which remain supported by the query after the candidates are removed, from improvements caused by copying candidate-only information into the reasoning trace.

\paragraph{Prompt conditions.}
For every query, we compare four query-side annotation conditions:
\begin{enumerate}
  \item \textbf{Input only}: the annotation model observes only the original multimodal
  query.
  \item \textbf{Input + positive}: the annotation model observes the query and its positive
  candidate.
  \item \textbf{Input + labeled positive/negative}: the annotation model observes the query, a candidate explicitly labeled as positive, and a candidate explicitly labeled as negative.
  \item \textbf{Input + blind candidate pair}: the annotation model observes the query and the same two candidates in a randomly shuffled A/B order, without being told which candidate is positive. It must first predict the positive candidate and then generate the reasoning trace.
\end{enumerate}

All conditions use the same task-specific instructions and output schema. The candidate-conditioned prompts instruct the annotation model to treat candidates as privileged attention cues rather than authoritative factual sources. 
Candidate information may be used to propose hypotheses about discriminative entities, attributes, relations, actions, spatial configurations, temporal events, or evidence types. 
However, the annotation model must re-inspect the query and retain only properties independently supported by the query. 
The final reasoning is instructed to be self-contained and must not mention candidates, positive/negative labels, comparisons, elimination, or A/B identifiers. 
The blind condition additionally outputs a separate top-level \texttt{choice} field, which is removed from the reasoning content used in the pairwise comparison.
Details of the used prompt can be seen in Appendix~\ref{app:prompt_templates}.

\paragraph{Reasoning generation.}
We use \texttt{GLM-4.1V-thinking} as the annotation model, served through a vLLM-compatible OpenAI API. We independently sample eight reasoning traces for each query under each of the four prompt conditions, producing $1{,}000 \times 4 \times 8 = 32{,}000$ annotations in total. Sampling uses temperature $1.0$, top-$p=0.95$, top-$k=20$, presence penalty $1.5$, and repetition penalty $1.0$. For video inputs, at most eight frames are supplied to the expert. The model returns a JSON object containing \texttt{reasoning} and \texttt{answer}; the blind condition additionally returns its predicted A/B choice.

The output-token limit was increased during generation to handle occasional empty responses from the serving backend. The limits used across generation segments were 4,096, 8,192, 80,000, and finally 16,384 tokens, while all prompts and sampling parameters remained unchanged. Only three successful generations reached 4,096 tokens and one reached 8,192 tokens, so truncation affected at most four of the 32,000 generated records. 

\paragraph{Blinded pairwise audit.}
We conduct a blinded, model-based pairwise audit at two scales. 
For the main trace-quality analysis, we evaluate all 1,000 queries in the probe dataset. For each query, we select one of the eight sampled traces from each prompt condition, using the same sampling index across conditions. Each candidate-conditioned trace is paired with the corresponding input-only trace, yielding $1{,}000 \times 3 = 3{,}000$ pairwise judgments over 4,000 selected traces. This audit supports the main comparison (Table \ref{tab:prompt-audit-overall}), candidate-leakage stratification (Table \ref{tab:prompt-audit-leakage}), domain-wise results (Table \ref{tab:prompt-audit-domains}), and blind-prediction correctness analysis (Table \ref{tab:prompt-audit-blind-correctness}) reported in the first four tables. Separately, we evaluate all eight samples per condition on an 81-query subset for the repeated-sampling analysis. Pairing each candidate-conditioned trace with the correspondingly indexed input-only trace yields $81 \times 3 \times 8 = 1{,}944$ pairwise judgments over 2,592 selected traces. This analysis supports the mean-of-eight, single-sample variability, and oracle best-of-eight results reported in Table \ref{tab:prompt-audit-scaling}.

For each pair, the input-only and candidate-conditioned traces are randomly assigned to the left and right positions. Pair order is also randomized. The reviewer has access to the complete multimodal query, positive candidate, hard negative, and both reasoning traces, but does not observe the prompt-condition identity or the hidden left/right mapping. The mapping is revealed only after all judgments are completed.

The reviewer answers the following question: \emph{Which trace would produce the better query embedding for retrieving the positive candidate while rejecting the hard negative after the candidates are removed?} A trace is preferred when it contains more query-supported discriminative information, omits fewer important query constraints, and contains fewer unsupported claims, generic statements, or candidate-only details.

\paragraph{Pairwise rating scale.}
Each comparison receives one of five ordinal ratings:
\[
r_i \in \{-2,-1,0,+1,+2\},
\]
where the sign is defined from the candidate-conditioned trace's perspective. A score of $+1$ indicates a small but meaningful advantage for the conditioned trace, and $+2$ indicates a clear advantage likely to affect positive-versus-negative ranking. Scores of $-1$ and $-2$ analogously favor the input-only trace, while $0$ denotes no meaningful retrieval-quality difference.

Let $N$ be the number of evaluated pairs for a prompt condition. We report:
\[
W=\sum_i \mathbb{I}[r_i>0],\quad
T=\sum_i \mathbb{I}[r_i=0],\quad
L=\sum_i \mathbb{I}[r_i<0].
\]
The \textbf{net win rate} measures the excess frequency of wins over losses:
\[
\operatorname{NetWin}=\frac{W-L}{N}.
\]
Unlike the ordinary win rate, this metric treats losses symmetrically and assigns ties zero contribution. Its range is $[-1,1]$.

The \textbf{mean margin} incorporates both preference direction and strength:
\[
\operatorname{Margin}=\frac{1}{N}\sum_{i=1}^{N}r_i,
\]
with range $[-2,2]$. A positive value favors candidate-conditioned reasoning. We compute 95\% confidence intervals for the mean margin using 10,000 bootstrap
replicates clustered by source query. Specifically, we first average the eight judgment margins for each query and then resample the 81 query-level means with replacement. This avoids treating repeated generations from the same query as independent observations.

\paragraph{Failure indicators.}
In addition to pairwise preferences, the reviewer independently marks three binary properties for each trace. \textbf{Candidate-only leakage} indicates that the trace contains a concrete entity, answer, number, date, quotation, category, or visual detail that is available from a candidate but cannot be independently inferred from the query. Its rate is
\[
\operatorname{LeakRate}
=\frac{1}{N}\sum_{i=1}^{N}\mathbb{I}[\text{conditioned trace }i\text{ leaks}].
\]
This is an operational measure of train--test information mismatch rather than benchmark contamination: such information is available to the posterior annotator but unavailable to the query-only model at deployment.

\textbf{Generic or vacuous} marks traces dominated by statements such as ``retrieve relevant information'' or ``focus on visual features'' without concrete query semantics. \textbf{Unsupported claim} marks factual, visual, OCR, spatial, or temporal assertions not justified by the query. Candidate specificity is counted as a genuine improvement only when it is independently supported by the query.

\paragraph{Repeated-sampling analyses.}
The primary result averages all eight stochastic samples per condition. We also report a \textbf{single-sample} estimate using the predefined sample with index zero. Finally, we report an \textbf{oracle best-of-eight} upper bound:
\[
\operatorname{OracleMargin}_s
=\max_{k\in\{1,\ldots,8\}} r_{s,k}.
\]
This oracle uses the pairwise audit score to select the best conditioned trace for each query and therefore should not be interpreted as a deployable result. It quantifies the potential benefit available to a sufficiently accurate verifier or reasoning-selection mechanism.

\paragraph{Blind verification analysis.}
For the blind condition, we additionally stratify generations by whether the expert correctly identifies the positive candidate. This is a conditional association rather than a causal estimate of verification: queries on which the expert predicts correctly may be intrinsically easier than those on which it predicts incorrectly. Within the 81-query audit set, 521 of 648 blind generations are prediction-correct and 127 are prediction-incorrect.

\begin{table}[t]
\centering
\small
\caption{Model-based pairwise audit of query-side reasoning on 1,000 MMEB-2
queries (v3-delimited prompts; random negatives). Each candidate-conditioned
trace is compared with the input-only trace for the same query and sample
index. W/T/L counts wins, ties, and losses; net win is $(W-L)/N$.
Margin is the mean signed preference score in $\{-2,-1,0,1,2\}$.
Intervals are query-level bootstrap 95\% confidence intervals for mean margin.
Leak denotes the fraction flagged for candidate-only information.}
\label{tab:prompt-audit-overall}
\resizebox{\linewidth}{!}{%
\begin{tabular}{lrrrrrr}
\toprule
Prompt condition & $N$ & W/T/L & Net win & Margin & 95\% CI & Leak \\
\midrule
Input + positive & 1000 & 445/393/162 & 0.283 & 0.451 & [0.388, 0.514] & 0.274 \\
Input + positive + negative (labeled) & 1000 & 406/416/178 & 0.228 & 0.385 & [0.322, 0.448] & 0.236 \\
Input + positive + negative (blind) & 1000 & 608/216/176 & 0.432 & 0.772 & [0.698, 0.845] & 0.796 \\
\bottomrule
\end{tabular}}
\end{table}

\begin{table}[t]
\centering
\small
\caption{Pairwise reasoning quality stratified by candidate-only information
leakage. Scores compare each trace with its matched input-only baseline.
These post-generation strata are descriptive and do not identify the causal
effect of leakage.}
\label{tab:prompt-audit-leakage}
\resizebox{\linewidth}{!}{%
\begin{tabular}{llrrrr}
\toprule
Prompt condition & Stratum & $N$ & W/T/L & Net win & Margin \\
\midrule
Input + positive & No leak & 726 & 238/367/121 & 0.161 & 0.249 \\
 & Leak & 274 & 207/26/41 & 0.606 & 0.985 \\
\addlinespace
Input + positive + negative (labeled) & No leak & 764 & 226/392/146 & 0.105 & 0.173 \\
 & Leak & 236 & 180/24/32 & 0.627 & 1.072 \\
\addlinespace
Input + positive + negative (blind) & No leak & 204 & 71/104/29 & 0.206 & 0.328 \\
 & Leak & 796 & 537/112/147 & 0.490 & 0.886 \\
\bottomrule
\end{tabular}}
\end{table}

\begin{table}[t]
\centering
\small
\caption{Pairwise reasoning-quality audit by MMEB-2 domain. All prompt
conditions include the query input and are compared against matched
input-only traces. Rates are reported as fractions.}
\label{tab:prompt-audit-domains}
\resizebox{\linewidth}{!}{%
\begin{tabular}{llrrrrr}
\toprule
Domain & Prompt condition & $N$ & W/T/L & Net win & Margin & Leak \\
\midrule
Image & Input + positive & 447 & 156/239/52 & 0.233 & 0.306 & 0.199 \\
 & Input + positive + negative (labeled) & 447 & 140/239/68 & 0.161 & 0.237 & 0.130 \\
 & Input + positive + negative (blind) & 447 & 196/150/101 & 0.213 & 0.391 & 0.725 \\
\addlinespace
Video & Input + positive & 220 & 101/87/32 & 0.314 & 0.473 & 0.164 \\
 & Input + positive + negative (labeled) & 220 & 94/87/39 & 0.250 & 0.400 & 0.136 \\
 & Input + positive + negative (blind) & 220 & 122/57/41 & 0.368 & 0.541 & 0.700 \\
\addlinespace
VisDoc & Input + positive & 333 & 188/67/78 & 0.330 & 0.631 & 0.447 \\
 & Input + positive + negative (labeled) & 333 & 172/90/71 & 0.303 & 0.574 & 0.444 \\
 & Input + positive + negative (blind) & 333 & 290/9/34 & 0.769 & 1.435 & 0.955 \\
\bottomrule
\end{tabular}}
\end{table}

\begin{table}[t]
\centering
\small
\caption{Blind-conditioned reasoning stratified by whether the expert
correctly selected the positive candidate. Unsupported denotes the fraction
of traces flagged for unsupported claims. Differences are descriptive:
correct and incorrect groups may differ in query difficulty and do not
establish a causal effect of verification.}
\label{tab:prompt-audit-blind-correctness}
\resizebox{\linewidth}{!}{%
\begin{tabular}{lrrrrrr}
\toprule
Blind prediction & $N$ & W/T/L & Net win & Margin & Leak & Unsupported \\
\midrule
Correct & 971 & 602/212/157 & 0.458 & 0.818 & 0.803 & 0.092 \\
Incorrect & 29 & 6/4/19 & -0.448 & -0.759 & 0.552 & 0.483 \\
\bottomrule
\end{tabular}}
\end{table}

\begin{table}[t]
\centering
\small
\caption{Repeated-sampling audit on 81 queries with eight samples per prompt
condition. Top: aggregate results over 648 paired comparisons per condition;
confidence intervals resample queries as clusters. Bottom: sample index 0
versus oracle-best-of-8, each with 81 comparisons per condition.
The oracle maximizes each query's paired margin against the input-only
trace at the corresponding sample index. It is a retrospective paired-score
upper bound, not an executable selection policy or an absolute-quality oracle.}
\label{tab:prompt-audit-scaling}
\resizebox{\linewidth}{!}{%
\begin{tabular}{lrrrrrr}
\toprule
Prompt condition & $N$ & W/T/L & Net win & Margin & 95\% CI & Leak \\
\midrule
Input + positive & 648 & 296/243/109 & 0.289 & 0.503 & [0.350, 0.660] & 0.250 \\
Input + positive + negative (labeled) & 648 & 265/262/121 & 0.222 & 0.395 & [0.248, 0.545] & 0.211 \\
Input + positive + negative (blind) & 648 & 441/117/90 & 0.542 & 0.937 & [0.739, 1.131] & 0.753 \\
\bottomrule
\end{tabular}}
\par\medskip
\resizebox{\linewidth}{!}{%
\begin{tabular}{llrr}
\toprule
Prompt condition & Selection & Net win & Margin \\
\midrule
Input + positive & Sample index 0 & 0.358 & 0.593 \\
 & Oracle-best-of-8 & 0.926 & 1.506 \\
\addlinespace
Input + positive + negative (labeled) & Sample index 0 & 0.099 & 0.259 \\
 & Oracle-best-of-8 & 0.889 & 1.383 \\
\addlinespace
Input + positive + negative (blind) & Sample index 0 & 0.568 & 0.988 \\
 & Oracle-best-of-8 & 0.951 & 1.580 \\
\bottomrule
\end{tabular}}
\end{table}

\subsection{Detailed RQ3 Experiment}
To check whether the transfer weights actually discriminate among posterior traces, we track their effective sample size during training of the Full model. For each query with at least one eligible posterior trace, we compute $\mathrm{ESS}=1/\sum_k w_k^2$ over its posterior traces, where a trace is eligible if its reward exceeds the mean reward of the prior group by more than $\max(0.05,\,4\sigma_{\text{prior}})$, and average over such queries at every step. As a reference we also report the mean number of eligible traces per query, which is the ESS that uniform weights would attain. The ESS rises from about 2.1 in the first 20 steps to about 3.6 in the last part of training, while the number of eligible traces grows to 6--7.5 per query. Throughout training the ESS stays at roughly half of this bound (mean ratio 0.53). The weights therefore do not spread evenly over the eligible traces but concentrate on a subset of them, so transfer is dominated by a few selected traces per query rather than by averaging over all traces that clear the eligibility threshold.
\label{app:rq3}
\begin{figure}[t]
\begin{center}
\includegraphics[width=1.0\textwidth]{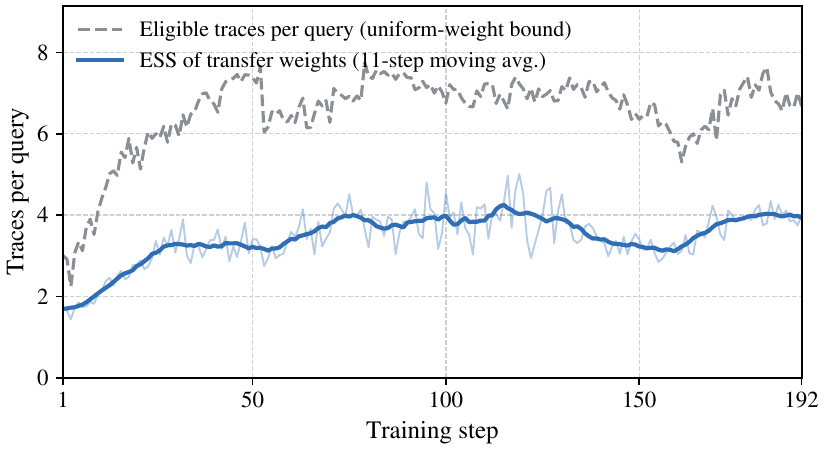}
\end{center}
\vspace{-10pt}
\caption{Effective sample size (ESS) of the transfer weights during training (Full).}
\label{fig:anlysis_q2&q3}
\vspace{-10pt}
\end{figure}

We measure how strongly the posterior traces depend on the privileged information $u$ through the signed
log-probability gap. For each posterior trace $z$ we compute the token-averaged difference
$\frac{1}{|z|}\sum_t \big[\log\pi_\theta(z_t \mid x,u,z_{<t}) - \log\pi_\theta(z_t \mid x,z_{<t})\big]$
between the posterior and prior prompts and average it over the batch. A lower gap means the trace is more likely under the prior prompt alone, i.e., it can be reproduced without access to $u$. We plot each dataset over its own block.
We compare Full with w/o TS, which disables transfer SFT. Full has a lower gap on all ten datasets, e.g., 0.30 versus 0.60 on CIRR, 0.33 versus 0.68 on Visual7W and 0.55 versus 0.83 on ColPali (averaged over each block).
Under w/o TS the gap also grows within several blocks, most visibly on LLaVA-Hound Caption and Visual7W, whereas under Full it stays flat or decreases. Transfer SFT thus keeps the posterior traces reachable by the prior, which is the property the transfer step relies on.
\begin{figure}[t]
\begin{center}
\includegraphics[width=1.0\textwidth]{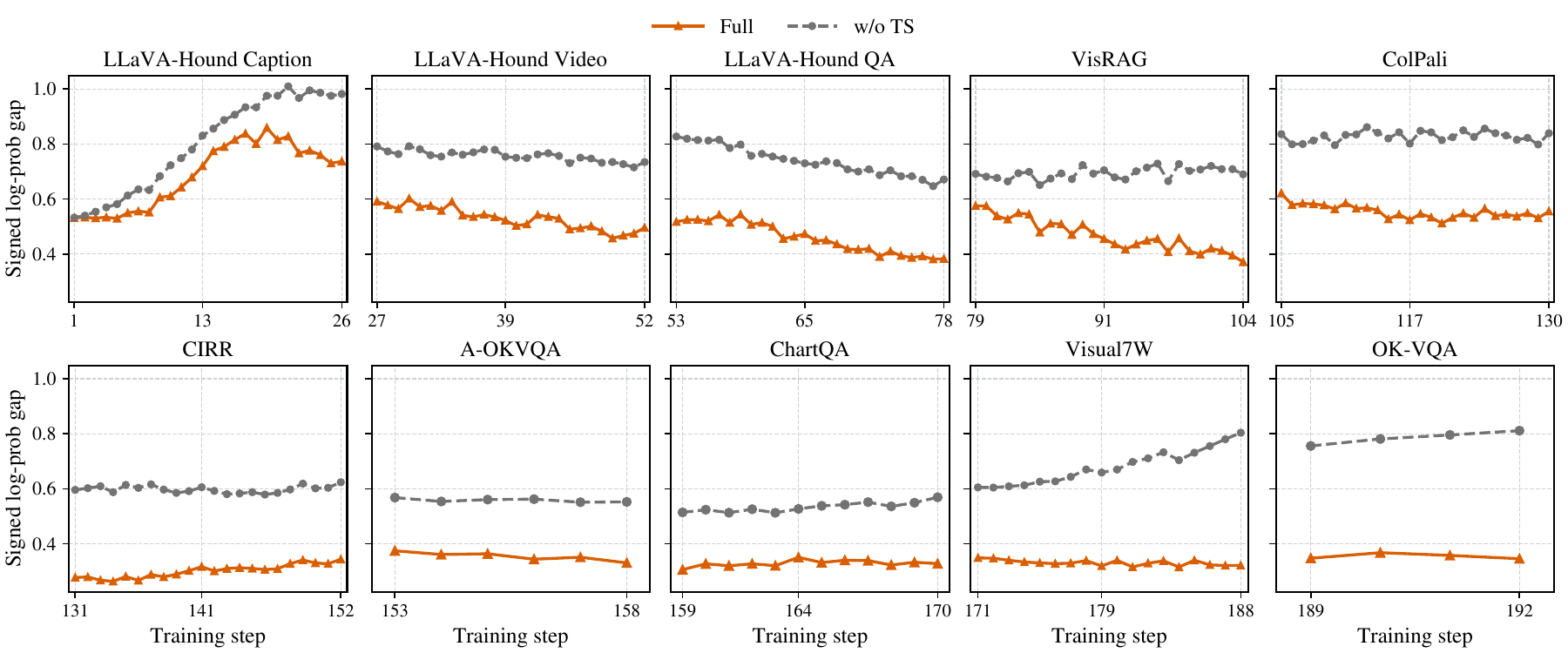}
\end{center}
\vspace{-10pt}
\caption{Signed log-probability gap between the posterior and prior prompts on posterior traces, per training subset.}
\label{fig:anlysis_q2&q3}
\vspace{-10pt}
\end{figure}

\end{document}